\documentclass[10pt,twocolumn,letterpaper]{article}

\usepackage{iccv}
\usepackage{times}
\usepackage{epsfig}
\usepackage{graphicx}
\usepackage{amsmath}
\usepackage{amssymb}

\usepackage[draft=true]{minted} 
\usepackage{bm}
\usepackage{xcolor}
\usepackage{multirow}
\usepackage{enumitem}
\usepackage{overpic}
\usepackage{etoolbox,xpatch}
\usepackage{style}

\usepackage[pagebackref=true,breaklinks=true,letterpaper=true,colorlinks,bookmarks=false]{hyperref}

\iccvfinalcopy 

\def\iccvPaperID{****} 

\ificcvfinal\fi

\begin{document}

\title{Deep Gaussian Processes for Few-Shot Segmentation}

\author{Joakim Johnander$^{1,2}$\quad Johan Edstedt$^1$\quad Martin Danelljan$^3$\quad Michael Felsberg$^1$\quad Fahad Shahbaz Khan$^{1,4}$ \and
  $^1$\small{Computer Vision Laboratory}\\ \small{Dept. of Electrical Engineering} \\ \small{Linköping University, Sweden} \and $^2$\small{Zenseact AB}\\ \small{Sweden} \and $^3$ \small{Computer Vision Lab}\\ \small{ETH Zurich, Switzerland} \and $^4$\small{Mohamed bin Zayed University of AI}\\ \small{UAE} \and
{\tt\small \{joakim.johnander,johan.edstedt,michael.felsberg\}@liu.se} \and {\tt\small {martin.danelljan@vision.ee.ethz.ch}} \and {\tt\small {fahad.khan@mbzuai.ac.ae}}}

\maketitle
\ificcvfinal\thispagestyle{empty}\fi

\begin{abstract}
Few-shot segmentation is a challenging task, requiring the extraction of a generalizable representation from only a few annotated samples, in order to segment novel query images. A common approach is to model each class with a single prototype. While conceptually simple, these methods suffer when the target appearance distribution is multi-modal or not linearly separable in feature space.

To tackle this issue, we propose a few-shot learner formulation based on Gaussian process (GP) regression. Through the expressivity of the GP, our approach is capable of modeling complex appearance distributions in the deep feature space. The GP provides a principled way of capturing uncertainty, which serves as another powerful cue for the final segmentation, obtained by a CNN decoder. We further exploit the end-to-end learning capabilities of our approach to learn the output space of the GP learner, ensuring a richer encoding of the segmentation mask.
We perform comprehensive experimental analysis of our few-shot learner formulation. Our approach sets a new state-of-the-art for 5-shot segmentation, with mIoU scores of $68.1$ and $49.8$ on PASCAL-5$^i$ and COCO-20$^i$, respectively.

\end{abstract}

\section{Introduction}
Few-shot segmentation (FSS)~\cite{Shaban2017} has received increased attention in recent years. The aim is to segment a query image given the support set, containing only a few annotated training samples of the class. Methods for FSS therefore need to extract information from the support set in order to segment the query image. A popular paradigm adopted by several recent works~\cite{Rakelly2018-2,ChiZhang2019,zhang2020sg,Nguyen2019} is \emph{meta-learning}. These works train a neural network directly for the FSS-task, where the network contains inductive priors in the form of purpose-specific few-shot learner modules.

\begin{figure}[t]
  \centering
  \includegraphics[width=\linewidth, trim={0.0cm 0.0cm 0.0cm 0.0cm}, clip]{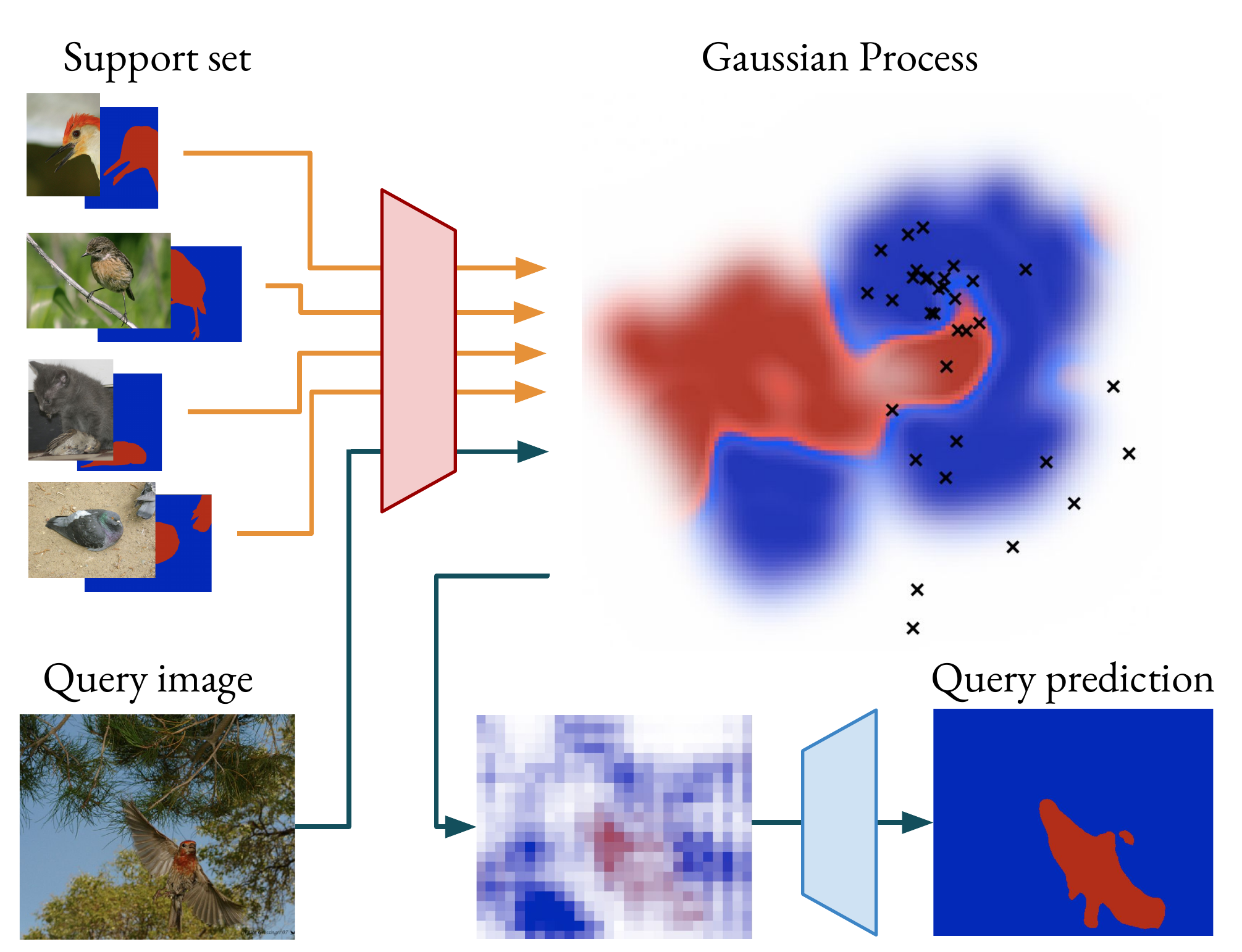}
  \caption{We use Gaussian processes to model the support set.  The learnt GP is visualized, with blue areas corresponding to background and red areas to foreground. White areas and areas with low contrast correspond to highly uncertain regions, given by the high variance predicted by the GP. Our approach leverages the flexible classification boundaries and uncertainties predicted by our GP few-shot learner.}
  \label{fig:illustration}
  \vspace{-0mm}
\end{figure}

Many recently proposed few-shot learner modules~\cite{Liu2020_DE,Tian2020GeneralizedFS,Wang2019_PA,zhang2020sg} extract a single prototype vector from the support set and and compare it to feature vectors extracted from the query image via a dot-product. In essence, this leads to a linear classifier and thus relies on the feature space being linearly separable over the classes. However, this is not necessarily the case. In fact, recent work~\cite{allen2019infinite} show that even when trained for such linear separability, the resulting feature space for \emph{unseen} classes is in many cases not linearly separable. A few works~\cite{Liu2020_PP,yang2020prototype} argue that even a single object may require different prototypes for different parts. In light of this, an attractive alternative would be to create a few-shot learning module that is more flexible, and able to model complex decision boundaries. However care must be taken to ensure that such a module does not overfit to the support set, and additionally it must be computationally tractable and fully differentiable. 

We explore using Gaussian process regression to model the support set and make predictions on the query set. In each episode, the support set is fed through an encoder to provide a set of features and corresponding values from the mask. Based on these, a GP is constructed. The GP probabilistically models the mapping between the features and the mask values and is flexible enough to act as an interpolator under certain assumptions on the noise~\cite{murphy2012machine}. During inference, the GP acts on query features and infers a probability distribution, the \emph{predictive posterior}, over the mask value for each feature. All calculations made during learning and inference of the GP are differentiable and rely on operations readily integrated as a module in a neural network.

Combining GPs with deep neural networks is appealing for several reasons. The neural network provides the GP with powerful visual features. In turn, the GP predictive posterior distribution gives a principled way of combining potentially highly correlated samples. Furthermore, the covariance in the predictive posterior provides additional uncertainty information. This information may be especially important in few-shot learning. As we are given only a few training examples, we are inevitably queried with unseen appearances. Whenever the GP is queried with unseen features, it will convey this information in the form of high covariance estimates. 

Since the Gaussian process is not limited to a one-dimensional output space, we go beyond predicting only a single mask value for each query feature. We encode the given support masks with a lightweight neural network. The resulting mask encoding is a multi-dimensional feature vector containing a richer representation of the mask. During inference, our Gaussian process few-shot learner makes a prediction on this output space. The predicted mask encodings, together with the uncertainty estimates, are fed to a subsequent decoder. This decoder is a neural network that combines these cues to produce a final segmentation.

\parsection{Contributions} Our main contributions are as follows.
\begin{enumerate}[label=\bfseries \roman*),topsep=0pt,itemsep=-1ex,partopsep=1ex,parsep=1ex]
    \item We propose a few-shot segmentation learner based on Gaussian processes. 
    \item We propose to learn the output space of the GP via a neural network that encodes the mask.
    \item We empirically show the benefits of the uncertainty information provided by the GP.
    \item We analyze how the approach scales with larger support sets on COCO-$20^i$.
    We demonstrate its effectiveness on the PASCAL-$5^i$ and COCO-$20^i$ benchmarks, outperforming prior works in the 5-shot setting.
\end{enumerate}

\section{Related Work}
\parsection{Prototype Based Few-Shot Segmentation}
Prototype-based methods aim to learn a \textit{single} representation of a semantic class based on the support set. Shaban \etal \cite{Shaban2017} proposed a method of predicting the weights of a linear classifier based on the support set, which was further built upon in later works \cite{Siam2019,Liu2020_DE}. Boudiaf \etal \cite{Boudiaf2020_repri} instead perform transductive inference at test-time to predict their classifier and achieve impressive results. Instead of learning the classifier directly, Rakelly \etal \cite{Rakelly2018-2} proposed to construct a global conditioning prototype from the support set and concatenate it to the query representation. This method proved successful, with a large number of subsequent works \cite{Dong2019,zhang2020sg,Wang2019_PA,Nguyen2019,ChiZhang2019,Liu2020_PP,azad2021texture,Liu2020_CR}. Wang \etal \cite{wang2021variational} additionally introduced a probabilistic perspective and model their prototype as a single Gaussian, which they can sample from at inference-time.
A fundamental limitation of the prototype based methods is the \textit{unimodality} assumption. Some works \cite{yang2020prototype,Liu2020_PP} try to circumvent this issue by clustering multiple prototypes to create a richer representation. However, clustering introduces extra hyperparameters as well as optimization difficulties. In contrast, our method is not restricted in any such sense, and only requires us to choose an appropriate kernel, which we will show is not a difficult task.

\parsection{Pointwise Few-Shot Segmentation}
Closer to our method are pointwise approaches. Instead of assuming a global representation of the target class, they aim to find  correspondences between the support and query set. Previous work has mostly focused on attention mechanisms \cite{Zhang2019_PG,yang2020new,hu2019attention}. Tian \etal~\cite{Tian2020} uses a mechanism similar to attention to generate a prior mask and then enrich it based on query features and a prototype. 
While the pointwise approach is much more flexible, we are left with the task of choosing how to optimally weigh the connections between the support and query set. Wang \etal \cite{Wang2020} approached this problem by additionally re-weighing the attention weights before the softmax operation to ensure a diverse support set. However, their approach is heuristic in nature and it is unclear in their framework how to optimally re-weigh the connections. In contrast to these works, our method provides a principled way of finding these correspondences, with the additional benefit of providing an uncertainty estimate which we will show greatly improves performance.

\begin{figure*}[t]
  \centering
  \includegraphics[width=\textwidth, trim={0.0cm 0.0cm 0.0cm 0.0cm}, clip]{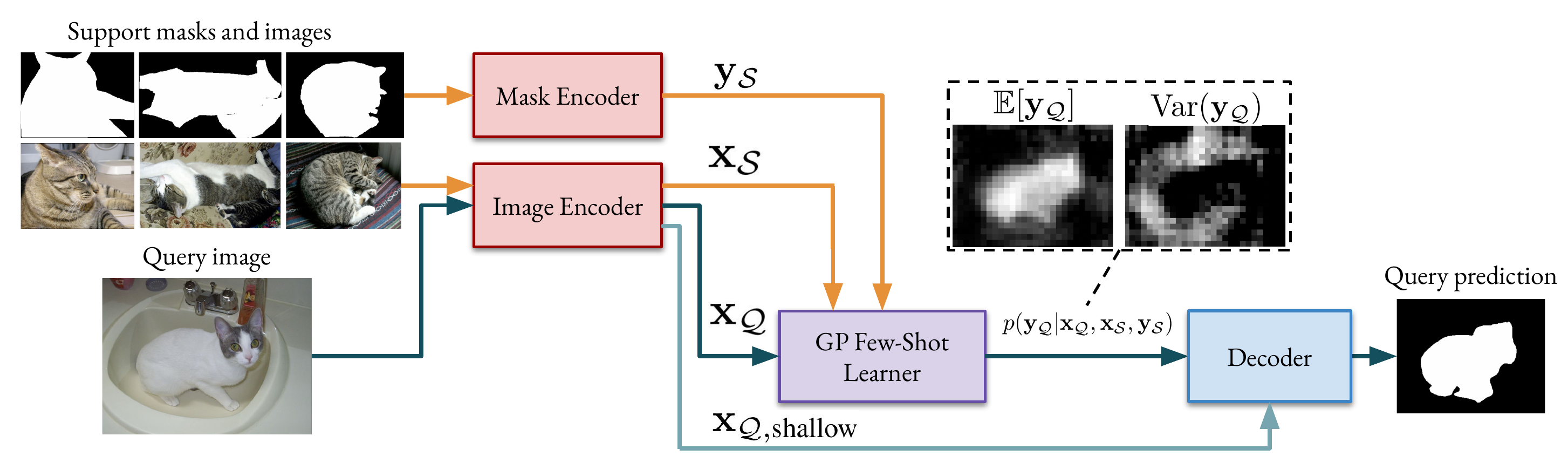}
  \caption{Overview of our approach. Support and query images are fed through an encoder to produce deep features $\mathbf{x}_\support$ and $\mathbf{x}_\query$ respectively. The support masks are fed through another encoder to produce $\mathbf{y}_\support$. Using Gaussian process regression, we infer the probability distribution of the query mask encodings $\mathbf{y}_\query$ given the support set and the query features (see equations \eqref{eq:predictive}-\eqref{eq:covariance}). We create a representation of this distribution and feed it through a decoder. The decoder then predicts a segmentation at the original resolution.}
  \label{fig:architecture}
\end{figure*}

\parsection{Combining GPs and Neural Networks} The combination of GPs and neural networks is not an entirely new idea. While earlier work focused on combining GPs and neural networks in the standard supervised setting \cite{Salakhutdinov2009,Wilson2016,Calandra2016}, there has recently been an increased interest in applying Gaussian processes in the context of \textit{meta-learning} \cite{Tossou2019} and \textit{few-shot classification} \cite{patacchiola2020bayesian,snell2021bayesian}. All these works employ the GP as a final \emph{output layer} in the network, and optimize either the predictive or marginal log likelihood directly. We go beyond this limitation and propose an internal GP model of the support features, where the posterior predictive distribution is used as input to a CNN decoder. Moreover, this allows us to learn the output space of the GP to further increase its expressive power.

\section{Method}
We propose an approach for few-shot segmentation where the modelling of the support set is based on Gaussian process regression. We briefly cover the few-shot segmentation problem and introduce the various components in Section~\ref{sec:overview}. We then discuss Gaussian process regression in Section~\ref{sec:gp}, followed by how we apply it to our task in Section~\ref{sec:metalearner}. In Section~\ref{sec:lwl} we extend the approach to encode the given support masks and in Section~\ref{sec:details} we go through architectural details.

\subsection{Overview}\label{sec:overview}
In few-shot segmentation, the aim is to learn a semantic segmentation method from a small collection of annotated images, called support set. For evaluation and training, we consider \emph{episodes}, each comprising a labelled support set and a collection of query images on which the segmentation itself is evaluated. This poses a major challenge compared to, for instance, semantic segmentation, since the model needs to extract information from the support set and compare it to the query image. Furthermore, the training and test data contain different semantic classes. This prohibits the model from \emph{cheating} and enforces learning from the support set.

An overview of our approach is given in Figure~\ref{fig:architecture}. First, the support and query images are fed through an image encoder to produce two sets of deep features, namely the support set $\{x_\support^i\}_i$ and the query set $\{x_\query^j\}_j$. Each feature $x^i\in\mathbb{R}^D$ corresponds to a spatial position in one of the images. We stack the support and query features as rows in a matrix to produce $\mathbf{x}_\support$ and $\mathbf{x}_\query$ respectively. The support masks are fed through another encoder to produce mask encodings $\{y_\support^i\}_i$ --- one per support feature. These are stacked into a vector $\mathbf{y}_\query$ to match the rows of $\mathbf{x}_\query$.
Now, the aim is to predict the mask encoding $y_\query^j$ corresponding to each query feature $x_\query^j$. The idea in this work is to use Gaussian processes to infer the conditional probability distribution $p(\mathbf{y}_\query|\mathbf{x}_\query, \mathbf{x}_\support, \mathbf{y}_\support)$, \ie the probability distribution of the mask encodings given the support set. A representation of this distribution is then fed to a decoder. The decoder combines the representation with shallow features extracted from the query image to predict a final segmentation.

\subsection{Gaussian Processes}\label{sec:gp}
Gaussian process regression is a method for supervised learning that, as we shall see, has several appealing properties for few-shot segmentation. We seek a function $f$ that relates the input to the output, \ie $f(x) = y$, given the support examples $\{x^i, y^i\}_i$. Gaussian processes infer a probability distribution over such functions. Using this distribution, we can then infer the conditional distribution of $y$ given previously unseen data $x$. In our case, the unseen data would be query features $x_\query^j$ and we would infer the distribution over the corresponding mask encoding $y_\query^j$. The idea in Gaussian process regression is to assume that the output is jointly Gaussian with some mean $\bm{\mu}$ and covariance $\bm{\Sigma}$. The covariance $\cov(y, y')$ between two outputs $y$ and $y'$ is defined by a covariance function or \emph{kernel} $\kappa$ that is evaluated at the two corresponding inputs. That is, $\cov(y, y') = \kappa(x, x')$. The kernel is a positive definite function and selected such that if its input features are similar, the corresponding outputs are also expected to be similar.

Gaussian processes provide a principled model $f$ with probabilistic output, containing both a point estimate as well as uncertainty information. GPs are non-parametric in the input space and stores all given input-output pairs. As such, they have the flexibility to model multi-modal input data as well as as complex decision boundaries~\cite{murphy2012machine}. The wrigglyness of the decision boundaries, or inversely, their smoothness, is directly controlled with the choice of kernel function $\kappa$. The features are utilized only within $\kappa$, with linear time complexity in the feature dimensionality. This makes GPs suitable for making predictions based on deep features. Importantly, inferring the distributions over unseen data is an entirely differentiable operation. A few-shot learner based on Gaussian processes is therefore readily integrated as a module within a deep neural network.

Going back to our few-shot learning problem, define $\mathbf{y}_\support$ and $\mathbf{y}_\query$ as the vectors of all support mask encodings $\mathbf{y}_\support$ and query mask encodings $\mathbf{y}_\query$ respectively. The former is observed and our aim is to predict a distribution over the latter. Here we have assumed that the mask encodings are one-dimensional for notational convenience. The GP assumes a joint distribution over the support and query samples given by,
\begin{align}
    \begin{pmatrix}\mathbf{y}_\support \\ \mathbf{y}_\query\end{pmatrix} \sim \mathcal{N} \left( \begin{pmatrix} \bm{\mu}_\support \\ \bm{\mu}_\query \end{pmatrix}, \begin{pmatrix} \mathbf{K}_{\support\support} & \mathbf{K}_{\support\query} \\ \mathbf{K}_{\support\query}\tp & \mathbf{K}_{\query\query} \end{pmatrix} \right)\enspace.
\end{align}
As most often the case, the prior means $\bm{\mu}_\support$ and $\bm{\mu}_\query$ are assumed to be zero. Instead, the covariances $\mathbf{K}_{\support\support} = \kappa(\mathbf{x}_\support, \mathbf{x}_\support)$, $\mathbf{K}_{\query\query} = \kappa(\mathbf{x}_\query, \mathbf{x}_\query)$, and $\mathbf{K}_{\support\query} = \kappa(\mathbf{x}_\support, \mathbf{x}_\query)$ determine the behavior of the GP. These are defined through the \emph{kernel} $\kappa:\mathbb{R}^D\times \mathbb{R}^D\rightarrow \mathbb{R}$. In our experiments, we adopt the commonly used \emph{squared exponential} (SE) kernel
\begin{align}
    \kappa(x^m, x^n) &= \sigma_f^2\exp(-\frac{1}{2\ell^2}\|x^m - x^n\|_2^2)\enspace,
\end{align}
with scale parameter $\sigma_f$ and length parameter $\ell$.

Next, we infer the posterior distribution of the query mask encodings $\mathbf{y}_\query$ using the support mask encodings $\mathbf{y}_\support$. That is, we calculate the parameters of the conditional
\begin{align}
    \mathbf{y}_\query|\mathbf{y}_\support, \mathbf{x}_\support, \mathbf{x}_\query &\sim \mathcal{N}(\bm{\mu}_{\query|\support},\bm{\Sigma}_{\query|\support})\enspace.
    \label{eq:predictive}
\end{align}
We assume that the measurements $\mathbf{y}_\support$ have been obtained with some additive i.i.d. Gaussian noise with variance $\sigma_y^2$. This corresponds to adding a scaled identity matrix $\sigma_y^2\mathbf{I}$ to the support covariance matrix $\mathbf{K}_{\support\support}$. Then, the parameters of the posterior are
\begin{align}
    \bm{\mu}_{\query|\support} &= \mathbf{K}_{\support\query}\tp(\mathbf{K}_{\support\support}+\sigma_y^2 \mathbf{I})^{-1} \mathbf{y}_\support\enspace,\label{eq:mean}\\
    \bm{\Sigma}_{\query|\support} &= \mathbf{K}_{\query\query} - \mathbf{K}_{\support\query}\tp(\mathbf{K}_{\support\support}+\sigma_y^2 \mathbf{I})^{-1} \mathbf{K}_{\support\query}\enspace.\label{eq:covariance}
\end{align}
For a derivation of these equations, we refer the reader to \cite{murphy2012machine}. With equations \eqref{eq:predictive}-\eqref{eq:covariance} we have calculated the distribution of the query mask encodings given our support set.

\subsection{Gaussian Processes for Meta Learning}\label{sec:metalearner}
Next, we integrate a GP meta learner in a few-shot segmentation framework. 
First, the $K$ support images $\{I_\support^k\}_k^K$ are fed through an image encoder $F^\text{feat}:\mathbb{R}^{3\times H\times W}\rightarrow\mathbb{R}^{D\times \frac{H}{16}\times \frac{W}{16}}$ to produce deep feature maps $X_\support^k$. The feature vectors in these deep feature maps are gathered to construct our support set $\mathbf{x}_\support$. The corresponding annotations, or support masks, are fed through the mask encoder $F^\text{mask}:\{0, 1\}^{H\times W}\rightarrow\mathbb{R}^{E\times\frac{H}{16}\times\frac{W}{16}}$ to produce corresponding mask encodings. These are gathered and stacked to produce the support mask encoding vector $\mathbf{y}_\support$. Note that, in contrast to several earlier works, we model both the background and the foreground. All features are utilized.

Using the same image encoder, we calculate the query features $\mathbf{x}_\query$. We obtain the posterior over the query mask encodings via \eqref{eq:predictive}-\eqref{eq:covariance} and therefore first compute the covariance matrix blocks $K_{\support\support}$, $K_{\support\query}$, and $K_{\query\query}$. We can then find the mean $\bm{\mu}_{\query|\support}$ and covariance $\bm{\Sigma}_{\query|\support}$ via \eqref{eq:mean} and \eqref{eq:covariance}. This involves the multiplication of a matrix inverse with another matrix. In practice, we use Cholesky decomposition and solve the resulting triangular systems of linear equations.

The posterior \eqref{eq:predictive} yields a joint probability distribution over the mask encodings. We are mostly interested in the marginal distribution of each component $y_\query^j$, which contains both the expected mask encoding and its uncertainty. We obtain this by taking the mean and the diagonal of the covariance, forming a representation of the distribution as
\begin{align}
    z^j &= [(\bm{\mu}_{\query|\support})_j, (\bm{\Sigma}_{\query|\support})_{j,j}]\enspace.
    \label{eq:z}
\end{align}
Here, $[\cdot]$ denotes concatenation and $j$ is used to index the mean and covariance. Note that, within the GP, the spatial order is irrelevant. The resulting query distribution is invariant over the order of the support set elements and it is equivariant over permutations of the query features. However, after constructing $z^j$, we arrange them to match the original feature map extracted from the image. The result is then fed into a decoder $F^\text{dec}:\mathcal{X}_{\query,\text{shallow}}\times\mathbb{R}^{(E + 1)\times\frac{H}{16}\times\frac{W}{16}}\rightarrow[0, 1]^{H\times W}$ that, with the help of shallow features $\mathbf{x}_{\query,\text{shallow}}\in\mathcal{X}_{\query,\text{shallow}}$, produces a final segmentation.

\subsection{Mask Encoder}\label{sec:lwl}
Thus far, we have assumed the mask encodings to be 1-dimensional values, \ie $E=1$. In the best of scenarios, the output from the GP module would be a low-resolution mask. We believe that the decoder would benefit from additional guidance. This could for instance be information about local shape or edges, anything that might help produce a fine segmentation mask. To that end, we feed the input mask through a lightweight neural network that outputs a multi-dimensional mask-encoding at the same resolution as the feature maps. This idea is not new. Recent work on Video Object Segmentation~\cite{bhat2020learning} shows the benefits of learning the output feature space, albeit with a different few-shot learning module.

Gaussian processes are able to model multi-dimensional output. For simplicity and computational reasons, we will assume the covariance function to be independent and isometric over the output dimensions. This means that the covariance becomes a block diagonal matrix. In practice, when the GP is used to infer $p(\mathbf{y}_\query|\mathbf{y}_\support, \mathbf{x}_\support, \mathbf{x}_\query)$ we find the mean via a series of matrix-vector multiplications \eqref{eq:mean}. If the mask encodings are multi-dimensional vectors we stack them as rows in a matrix. The columns represent different dimensions, and the resulting mean is a matrix of the same size. Essentially, \eqref{eq:mean} is calculated once per dimension of the mask encoding. The resulting representation $z^j$ then contains a multi-dimensional mean and a single value for the variance. Note that while having a general covariance in the multi-dimensional case is possible, it quickly becomes computationally intractable.

\subsection{Architecture Details}\label{sec:details}
Following previous works we adopt a ResNet-50~\cite{he2016deep} backbone pre-trained on ImageNet~\cite{ILSVRC15} as our image encoder. We dilate its last residual module by a factor of 2 and remove the terminal average pooling to provide features with stride 16. In addition, we place a single convolutional projection layer that reduces the the 2048-dimensional feature map down to 512 dimensions. As mask encoder we adopt the publicly available LWL-encoder~\cite{bhat2020learning}. For the Gaussian process few-shot learner, we downsample the support feature maps and annotation encodings with a factor of 2, yielding a total stride of 32 compared to the original images. We use $\sigma_y^2=0.01$, $\sigma_f^2=1$, and $\ell^2 = \sqrt{512}$. As our decoder we adopt the segmentation decoder DFN proposed in \cite{Yu2018dfn}. We do not add the border network of DFN and we set it to upsample with the shallow feature maps output by the first and second ResNet modules. We implement the approach in PyTorch~\cite{pytorch} and code will be made available at publication.

\begin{table*}[t]
    \centering
    \begin{tabular}{c | c | c | c c c c c | c c c c c }
        \thickhline
        \multirow{2}{*}{Method}& \multirow{2}{*}{Backbone}   & \multirow{2}{*}{Size}          &        &        &1-Shot  &        &        &        &        & 5-Shot &        & \\
                          &&& F-0 & F-1 & F-2 & F-3 &  Mean  & F-0 & F-1 & F-2 & F-3 & Mean \\
        \thickhline
        OSLSM \cite{Shaban2017}&  VGG16& Orig       & 33.6      & 55.3      & 40.9      & 33.5       & 40.8       & 35.9       & 58.1       & 42.7      &  39.1      &  43.9\\
        PANet\cite{Wang2019_PA}&  VGG16 & 417       & 42.3      & 58.0      & 51.1      & 41.2       & 48.1       & 51.8       & 64.6       & 59.8      &  46.5      &  55.7\\
        CANet \cite{ChiZhang2019}&  RN50  &  Orig    & 52.5      & 65.9      & 51.3      & 51.9       & 55.4       & 55.5       & 67.8       & 51.9      &  53.2      &  57.1\\
        RPMM \cite{yang2020prototype}&  RN50 & ? &  55.2    & 66.9      & 52.6      & 50.7      & 56.3       & 56.3       & 67.3       & 54.5       & 51.0      &  57.3\\
        PGNet$^{MS}$ \cite{Zhang2019_PG} & RN50 & Orig &  56.0        & 66.9      & 50.6      & 50.4      & 56.0        & 57.7       & 68.7       & 52.9       & 54.6      & 58.5\\
        CRNet\cite{Liu2020_CR}&  RN50& Orig        & -      & -      & -      & -       & 55.7       &-       &-       & -      &  -      &  58.8\\
        FWB \cite{Nguyen2019}&  RN101    & Orig   & 51.3      & 64.5      & 56.7      & 52.2       & 56.2       & 54.8       & 67.4       & 62.2      &  55.3      &  60.0\\
        VPI \cite{wang2021variational} & RN101& ?        & 53.4      & 65.6      & 57.3      & 52.9       & 57.3       &55.8       &67.5       & 62.6      &  55.7      &  60.4\\
        DENet \cite{Liu2020_DE}&  RN50 & 321      & 55.7      & \textbf{69.7}      & \textbf{63.6}      & 51.3       & 60.1       & 54.7       & 71.0       & 64.5      &  51.6      &  60.5\\
        DAN$^{MS}$ \cite{Wang2020}&RN101&   ?       & 54.7      & 68.6      & 57.8      &51.6       &58.2       &57.9       &69.0       & 60.1      & 54.9        & 60.5\\
        LTM \cite{yang2020new}& RN50 & ? &  52.8 & 69.6 & 53.2 & 52.3 & 57.0  & 57.9 & 69.9 & 56.9 & 57.5 & 60.6\\
        PFENet \cite{Tian2020}&  RN50 & Orig       & \textbf{61.7}      & 69.5      & 55.4      & \textbf{56.3}       & \textbf{60.8}       & 63.1       & 70.7       & 55.8      &  57.9      &  61.9\\
        PPNet$^\dag$ \cite{Liu2020_PP}&RN50      & 417   & 47.8      & 58.8      & 53.8      & 45.6       & 51.5       & 58.4       & 67.8       & 64.9      & 56.7        & 62.0\\

        CAPL \cite{Tian2020GeneralizedFS}&  RN101& Orig       & -      & -      & -      & -       &57.4       &-       &-       & -      &  -      &  63.4\\
        RePri$^\ddag$ \cite{Boudiaf2020_repri}&RN50 & 417         & 59.8      & 68.3      & 62.1      & 48.5       & 59.7       & 64.6       &\textbf{71.4}       & 71.1      & 59.3       & 66.6\\

        \hline\textbf{Ours}     & RN50      & Orig & 50.5 & 64.9       & 54.6      & 52.0       & 55.5& \textbf{66.8}      & 70.7      & \textbf{71.6}       & \textbf{63.2}       &\textbf{68.1}\\
        
    \thickhline
    \end{tabular}
    \caption{Performance on PASCAL-$5^i$ compared to the state-of-the-art (mIoU, higher is better). \emph{Size} is the resolution used to calculate mIoU at test-time. F-X is performance on fold X of the benchmark.
    $^{MS}$Multi-scale inference
    $^\dag$We report results without additional unlabeled data during test-time
    $^\ddag$Transductive inference instead of meta-learning}
    \label{tab:pascal-sota}
\end{table*}

\begin{table*}[t]
    \centering
    \begin{tabular}{c | c | c | c c c c c | c c c c c }
        \thickhline
        \multirow{2}{*}{Method}& \multirow{2}{*}{Backbone} & \multirow{2}{*}{Size}         &        &        &1-Shot  &        &        &        &        & 5-Shot &        & \\
                          &&& F-0 & F-1 & F-2 & F-3 &  Mean  & F-0 & F-1 & F-2 & F-3 & Mean \\
        \thickhline
        FWB \cite{Nguyen2019}&  RN101 & Orig        & 17.0      & 18.0      & 21.0      & 28.9       & 21.2       & 19.1       & 21.5       &    23.9      &  30.1      &  23.7\\
        VPI \cite{wang2021variational}&  RN101 & ?        & - & - & - & - & 23.4 & - & - & - & - & 27.8\\
        DAN$^{MS}$ \cite{Wang2020} &RN101  &  ?      & -      & -      & -      &-        & 24.4       & -       & -       & -      & -       & 29.6\\
        RPMM \cite{yang2020prototype} &RN50 & ? &  29.5      & 36.8      & 28.9      & 27.0      & 30.6        & 33.8       & 42.0       & 33.0       & 33.3      & 35.5\\
        PPNet$^\dag$ \cite{Liu2020_PP} & RN50    & 417      & 34.5      & 25.4      & 24.3      &18.6       &25.7       &48.3       &30.9       & 35.7      & 30.2       & 36.2\\
        PFENet \cite{Tian2020}&  RN101 & Orig        & 34.3      & 33.0      & 32.3      & 30.1       & 32.4       & 38.5       & 38.6       & 38.2      &  34.3      &  37.4\\
        PGNet$^*$ \cite{Zhang2019_PG}&  RN50 & 321      & 39.5      & 39.7      & 33.9      & 33.5       & 36.7       & 42.4       & 38.9       & 32.4      &  36.5      &  37.5\\
        CAPL \cite{Tian2020GeneralizedFS}&  RN101& Orig       & -      & -      & -      & -       & 34.5       & -       & -       & -      &  -      &  41.1\\        
        RePri$^\ddag$ \cite{Boudiaf2020_repri}& RN50 & 417 & 32.0      & 38.7      & 32.7      & 33.1       & 34.1       & 39.3       & 45.4       & 39.7      & 41.8    & 41.6 \\
        CANet$^*$ \cite{ChiZhang2019}&  RN50 & 321      & 42.2      & 42.7      & 37.6      & \textbf{40.9}       & 40.8       & 44.7       & 43.0       & 37.5      &  42.7      &  42.0\\
        PFENet \cite{Tian2020}&  RN101 & 473    & 36.8      & 41.8      & 38.7      & 36.7       & 38.5       & 40.4       & 46.8       & 43.2      &  40.5      &  42.7\\
        DENet \cite{Liu2020_DE}&  RN50& 321 &\textbf{42.9}      & \textbf{45.8}      & \textbf{42.2}      & 40.2       & \textbf{42.8}       & 45.4       & 44.9       & 41.6      &  40.3      &  43.0\\
        
        \hline\textbf{Ours}     & RN50      & Orig      & 37.1      & 40.6       & 38.6        & 37.7       & 38.5       & \textbf{48.4}      & \textbf{53.4}       & \textbf{49.5}       & \textbf{48.0}       & \textbf{49.8}\\
        
    \thickhline
    \end{tabular}
    \caption{Performance on COCO-$20^i$ compared to the state-of-the-art (mIoU, higher is better). \emph{Size} is the resolution used to calculate mIoU at test-time. F-X is performance on fold X of the benchmark.
    $^*$Re-implementation by \cite{Liu2020_DE}
    $^{MS}$Multi-scale inference
    $^\dag$We report results without additional unlabeled data during test-time
    $^\ddag$Transductive inference instead of meta-learning}
    \label{tab:coco-sota}
\end{table*}

\section{Experiments}
We conduct experiments on the FSS benchmarks PASCAL-$5^i$~\cite{Shaban2017} and COCO-$20^i$~\cite{Nguyen2019}. First, we describe our experimental setup in detail. Second, we perform a thorough state-of-the-art comparison, including an experiment showing how well the GP learner scales with additional shots. This is followed by an ablation study containing both a quantitative and a qualitative comparison. Last, we analyze the choice of kernel function $\kappa$.

\begin{figure}[t]
  \centering
  \includegraphics[width=\linewidth, trim={0.0cm 0.0cm 0.0cm 0.0cm}, clip]{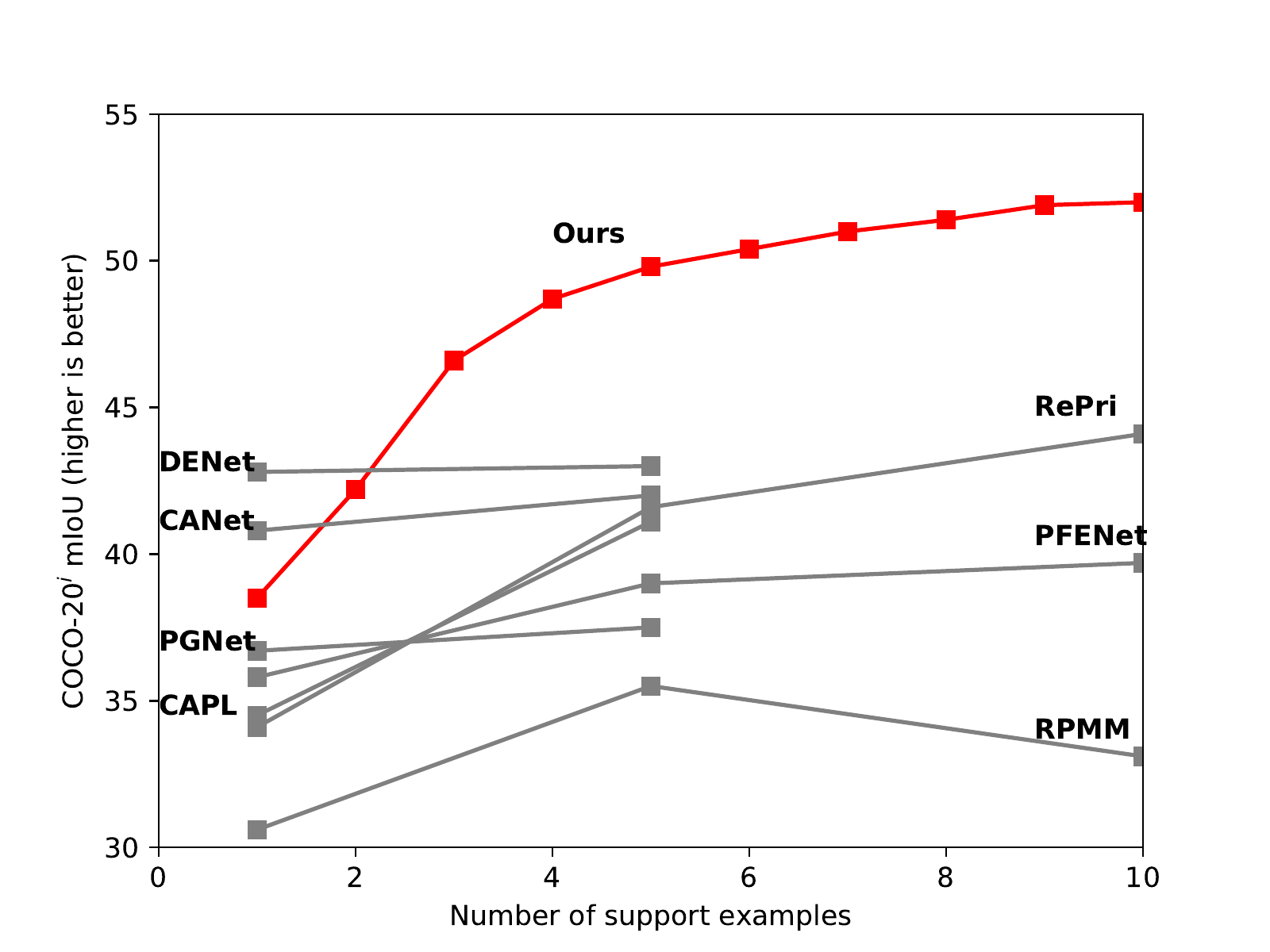}
  \caption{Performance of the proposed approach on COCO-$20^i$ for different number of shots, compared to the state-of-the-art (mIoU, higher is better). 10-shot numbers obtained from Boudiaf \etal~\cite{Boudiaf2020_repri}. As the number of shots increase from 1 to 10, the performance of the proposed approach monotonically increases. Already at 3 shots, the proposed approach outperforms the state-of-the-art in the 5-shot setting.}
  \label{fig:kshot_performance}
\end{figure}

\subsection{Experimental Setup}\label{sec:exp_setup}
\parsection{Datasets}
We use the PASCAL-$5^i$~\cite{Shaban2017} and COCO-$20^i$~\cite{Nguyen2019} benchmarks for our experiments. PASCAL-$5^i$ is composed of PASCAL VOC 2012~\cite{Everingham2010} with additional SBD~\cite{Hariharan2011} annotations. The dataset is split into 4 folds à 5 classes. COCO-$20^i$~\cite{Nguyen2019} is built from MS-COCO~\cite{Lin2014} and is in a similar fashion split into 4 folds à 20 classes. Both benchmarks measure performance via 4-fold cross-validation. Three folds are used for training and the last fold held out for testing. Training is done on the respective training sets and testing on the respective validation sets. Thus, during the cross-validation, both the images \emph{and} classes differ between the training and test sets.

\parsection{Training Details}
We train our models for 20000 and 40000 iterations for PASCAL-$5^i$ and COCO-$20^i$ respectively. We sample 8 episodes per batch and use the Adam~\cite{Kingma2015adam} optimizer. We use a learning rate of $10^{-5}$ that is decayed with a factor of $0.3$ halfway through training. During training, we freeze the batch normalization layers in our image encoder. We adopt the episode sampling and data augmentation used in \cite{Tian2020}. First, all images are resized to fit in a predefined window. We use a window of size 448 for PASCAL and 512 for COCO. Then, the sample is zero-padded and augmented. The augmentation comprises scale jitter in the range $\{0.9, 1.1\}$, rotations in the range $\{-10, 10\}$, Gaussian blur, and random horizontal flips. 

\parsection{Evaluation}
We evaluate our approach on each fold using 5000 and 20000 episodes respectively, for PASCAL-$5^i$ and COCO-$20^i$. This follows the work of Tian \etal~\cite{Tian2020} in which it is observed that the procedure employed by many prior works, using only 1000 episodes, yields fairly high variance. Performance is measured in terms of mean Intersection over Union (mIoU). First, the Intersection over Union (IoU) is calculated per class over all episodes in a fold. The mIoU is then found by averaging the IoU over the classes. Like the original work on FSS by Shaban \etal~\cite{Shaban2017}, we calculate the IoU on the original resolution.

\subsection{State-of-the-Art Comparison}\label{sec:sota}
We compare the proposed approach to the state-of-the-art on PASCAL-$5^i$ and COCO-$20^i$ and show the results in Table~\ref{tab:pascal-sota} and Table~\ref{tab:coco-sota} respectively. Following prior works, we report results given a single support example, \emph{1-shot}, and given five support examples, \emph{5-shot}. On PASCAL-$5^i$, our approach obtains a respectable 1-shot performance of 55.5 mIoU. In the 5-shot setting, our approach outperforms the state-of-the-art with an mIoU of 68.1. This corresponds to an absolute mIoU gain of 1.5 compared to previous best method, RePri~\cite{Boudiaf2020_repri}, and a 4.7 gain compared to the previous best meta-learning method, CAPL~\cite{Tian2020GeneralizedFS}.

On COCO-$20^i$, our approach obtains a competitive 1-shot performance of 38.5 mIoU. In the 5-shot setting, the top performing approaches~\cite{Tian2020GeneralizedFS,Boudiaf2020_repri,ChiZhang2019,Tian2020,Liu2020_DE} obtain 41-43 mIoU. These include prototype based approaches~\cite{Tian2020GeneralizedFS, ChiZhang2019,Liu2020_DE}, a point-based approach~\cite{Tian2020}, and an approach not based on meta-learning~\cite{Boudiaf2020_repri}. Our approach obtains 49.8 mIoU, an absolute increase of 6.8 mIoU. This clearly demonstrates the effectiveness of the GP-based few-shot learner.

An important property for few-shot segmentation is to scale with more data. We therefore analyze the effectiveness of the proposed approach as the number of shots increases. The results are shown in Figure~\ref{fig:kshot_performance}. Our approach is trained for 5 shots, except for the 1-shot result in which case we train for 1 shot. The performance of the proposed approach monotonically increases as the number of shots increases. At three shots, the GP-based few-shot learner significantly outperforms the state-of-the-art, and the performance keeps increasing all the way to 10 shots. While DENet~\cite{Liu2020_DE} and CANet~\cite{ChiZhang2019} provides high 1-shot performance, the performance increases only slightly when instead given 5 support examples. Both these methods rely on single prototypes. We believe that this representation struggles to model multiple support images. The GP, in contrast, is expressive enough to accurately model the given support set.

\subsection{Ablation Study}\label{sec:ablation} 
\begin{figure*}
    \hspace{.28\columnwidth}Support Set $\support$ \hspace{.28\columnwidth}\hspace{.04\columnwidth}Ground Truth Query \hspace{.04\columnwidth}\hspace{.12\columnwidth} Baseline \hspace{.12\columnwidth}\hspace{.15\columnwidth} \textbf{Ours}
    
    \begin{overpic}[width=.4\columnwidth]{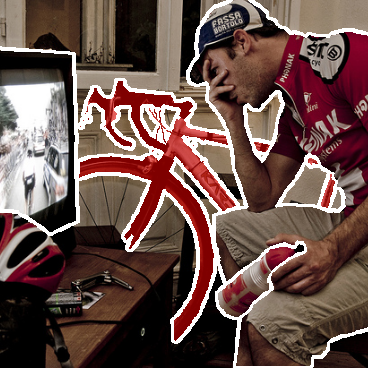}
    \put(100,0){\includegraphics[width=.2\columnwidth]{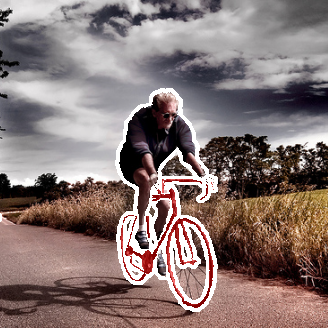}}
    \put(100,50){\includegraphics[width=.2\columnwidth]{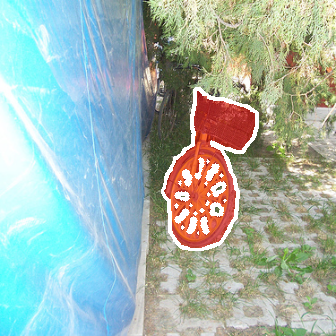}}
    \put(150,0){\includegraphics[width=.2\columnwidth]{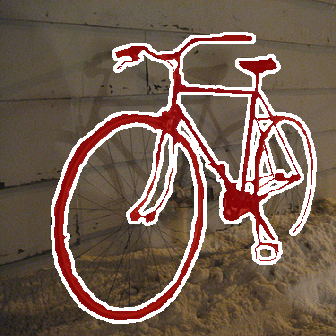}}
    \put(150,50){\includegraphics[width=.2\columnwidth]{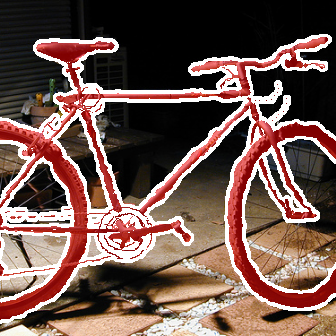}}
    \end{overpic}\hspace{.41\columnwidth}
    \includegraphics[width=.4\columnwidth]{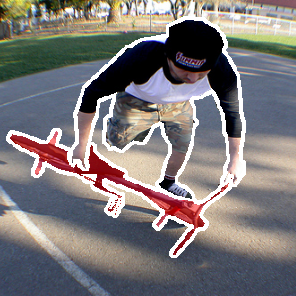}\hspace{0.01\columnwidth}
    \includegraphics[width=.4\columnwidth]{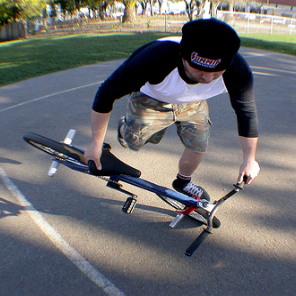}\hspace{0.01\columnwidth}
    \includegraphics[width=.4\columnwidth]{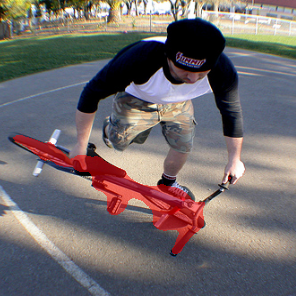}
    
    \vspace{0.01\columnwidth}
    
    \begin{overpic}[width=.4\columnwidth]{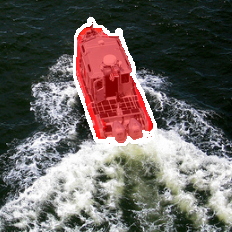}
    \put(100,0){\includegraphics[width=.2\columnwidth]{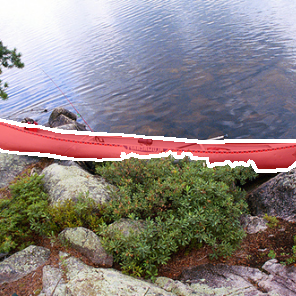}}
    \put(100,50){\includegraphics[width=.2\columnwidth]{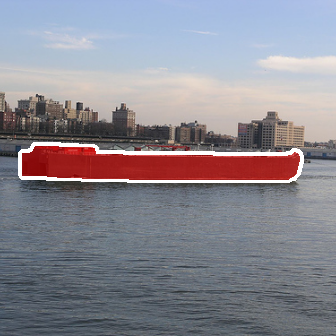}}
    \put(150,0){\includegraphics[width=.2\columnwidth]{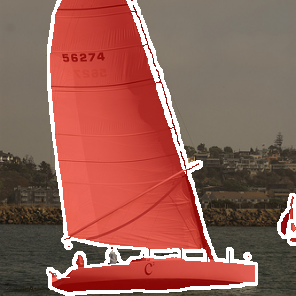}}
    \put(150,50){\includegraphics[width=.2\columnwidth]{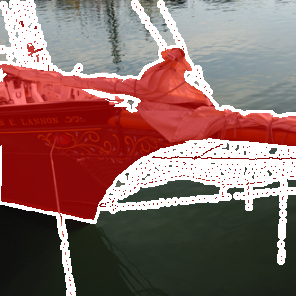}}
    \end{overpic}\hspace{.41\columnwidth}
    \includegraphics[width=.4\columnwidth]{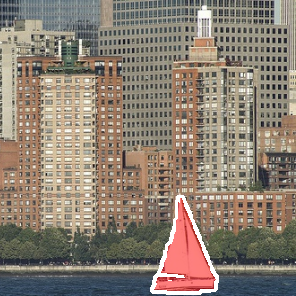}\hspace{0.01\columnwidth}
    \includegraphics[width=.4\columnwidth]{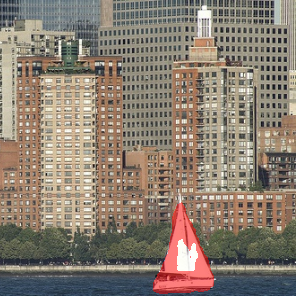}\hspace{0.01\columnwidth}
    \includegraphics[width=.4\columnwidth]{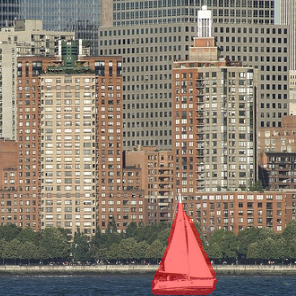}
    
    \vspace{0.01\columnwidth}

    \begin{overpic}[width=.4\columnwidth]{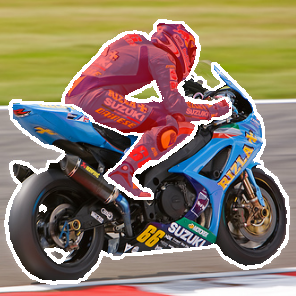}
    \put(100,0){\includegraphics[width=.2\columnwidth]{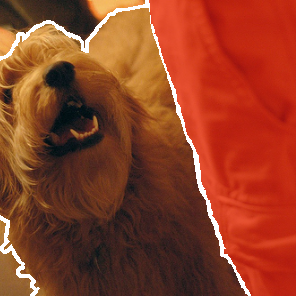}}
    \put(100,50){\includegraphics[width=.2\columnwidth]{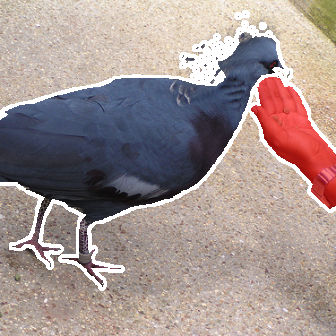}}
    \put(150,0){\includegraphics[width=.2\columnwidth]{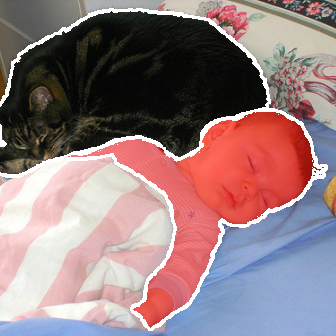}}
    \put(150,50){\includegraphics[width=.2\columnwidth]{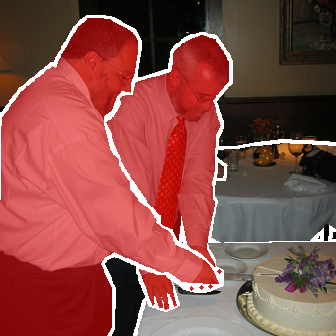}}
    \end{overpic}\hspace{.41\columnwidth}
    \includegraphics[width=.4\columnwidth]{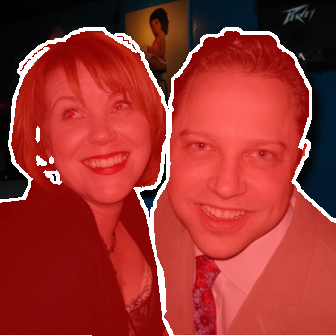}\hspace{0.01\columnwidth}
    \includegraphics[width=.4\columnwidth]{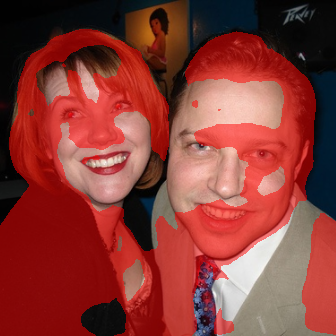}\hspace{0.01\columnwidth}
    \includegraphics[width=.4\columnwidth]{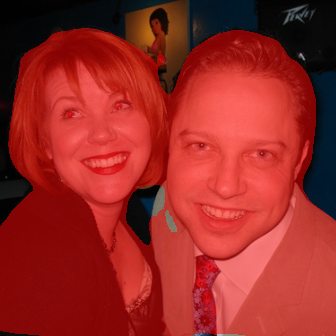}
    
    \vspace{0.01\columnwidth}
    
    \begin{overpic}[width=.4\columnwidth]{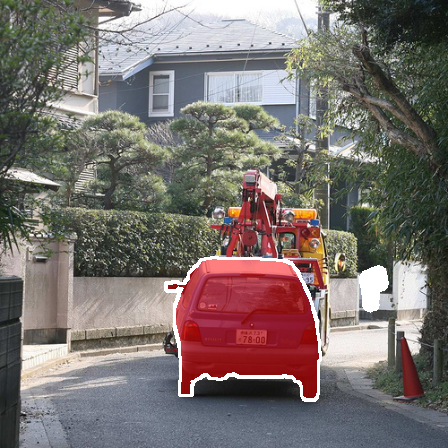}
    \put(100,0){\includegraphics[width=.2\columnwidth]{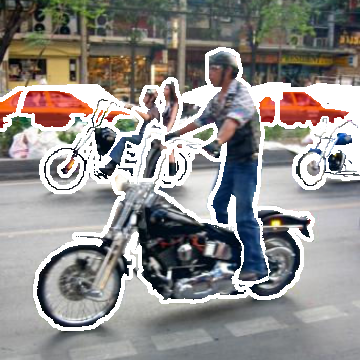}}
    \put(100,50){\includegraphics[width=.2\columnwidth]{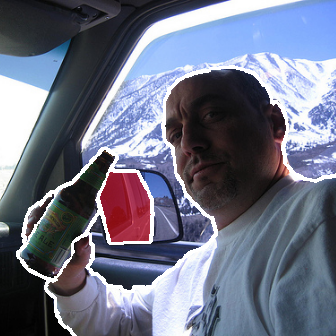}}
    \put(150,0){\includegraphics[width=.2\columnwidth]{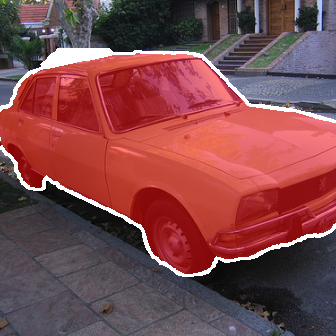}}
    \put(150,50){\includegraphics[width=.2\columnwidth]{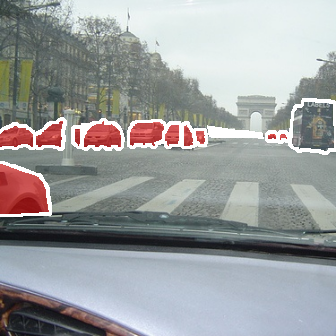}}
    \end{overpic}\hspace{.415\columnwidth}
    \includegraphics[width=.4\columnwidth]{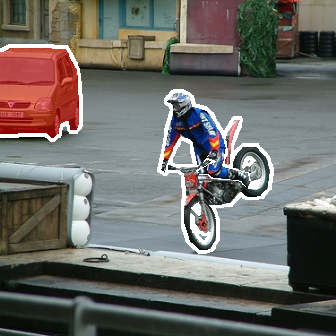}\hspace{0.01\columnwidth}
    \includegraphics[width=.4\columnwidth]{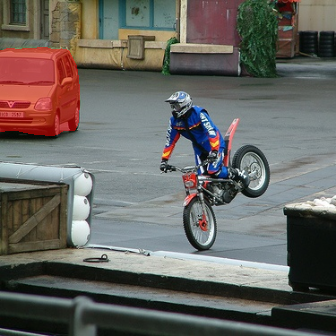}\hspace{0.01\columnwidth}
    \includegraphics[width=.4\columnwidth]{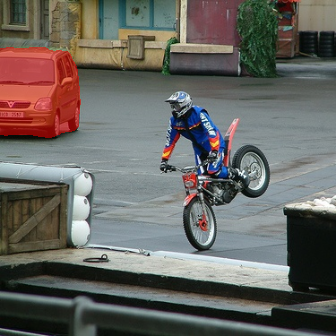}
    
    \vspace{0.01\columnwidth}
    
    \begin{overpic}[width=.4\columnwidth]{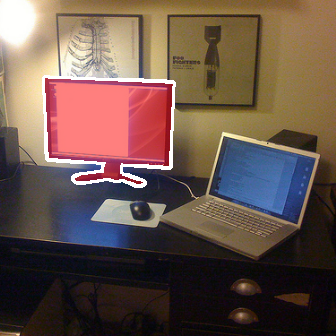}
    \put(100,0){\includegraphics[width=.2\columnwidth]{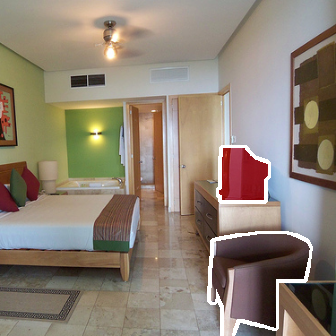}}
    \put(100,50){\includegraphics[width=.2\columnwidth]{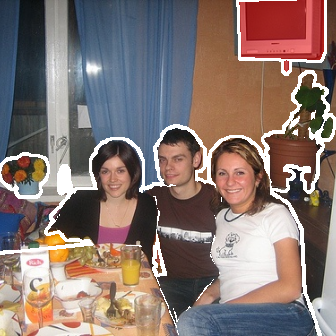}}
    \put(150,0){\includegraphics[width=.2\columnwidth]{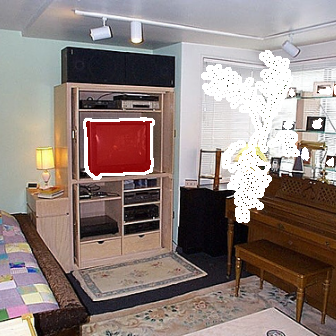}}
    \put(150,50){\includegraphics[width=.2\columnwidth]{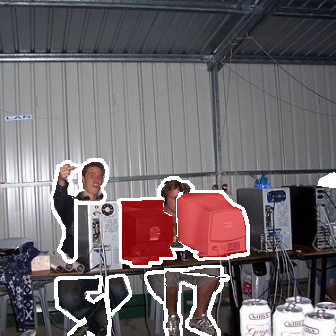}}
    \end{overpic}\hspace{.41\columnwidth}
    \includegraphics[width=.4\columnwidth]{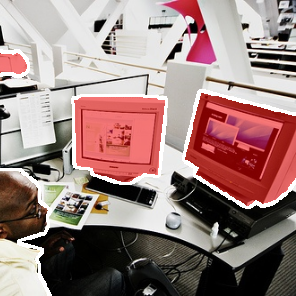}\hspace{0.01\columnwidth}
    \includegraphics[width=.4\columnwidth]{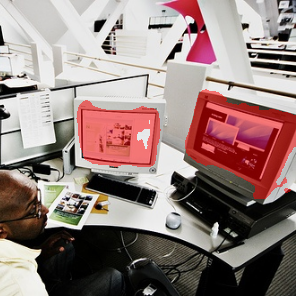}\hspace{0.01\columnwidth}
    \includegraphics[width=.4\columnwidth]{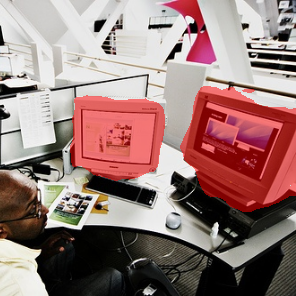}\vspace{0.01\columnwidth}

    \begin{overpic}[width=.4\columnwidth]{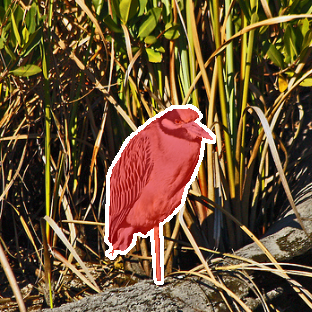}
    \put(100,0){\includegraphics[width=.2\columnwidth]{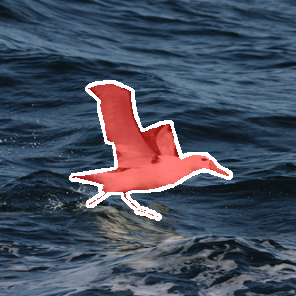}}
    \put(100,50){\includegraphics[width=.2\columnwidth]{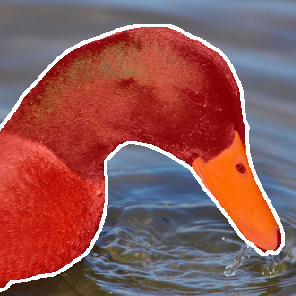}}
    \put(150,0){\includegraphics[width=.2\columnwidth]{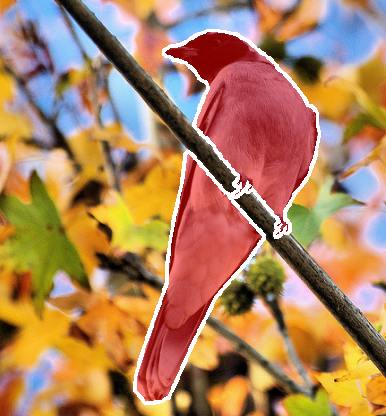}}
    \put(150,50){\includegraphics[width=.2\columnwidth]{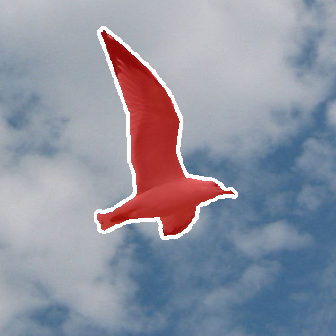}}
    \end{overpic}\hspace{.41\columnwidth}
    \includegraphics[width=.4\columnwidth]{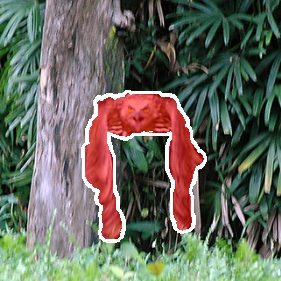}\hspace{0.01\columnwidth}
    \includegraphics[width=.4\columnwidth]{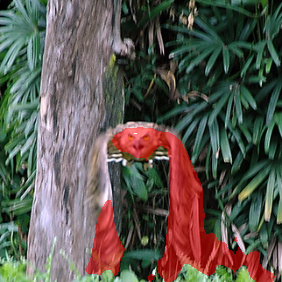}\hspace{0.01\columnwidth}
    \includegraphics[width=.4\columnwidth]{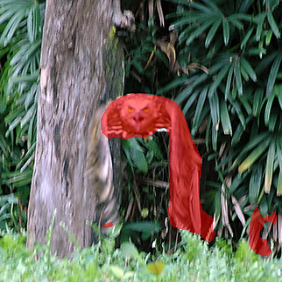}
\caption{Challenging episodes in the 5-shot setting from the PASCAL-$5^i$ benchmark. We show the five support images with annotations overlaid (left), the query image and annotation (center), predictions made by our baseline method (right), and predictions made by our final model (rightmost). Note that the images are cropped for aesthetic reasons.}
\label{fig:qualitative}
\end{figure*}

Here, we analyze the impact of the key components in our proposed architecture. Quantitative results are shown in Table \ref{tab:ablation}, while qualitative results are shown in Figure \ref{fig:qualitative}.

\parsection{Mean}First, we alter equation \eqref{eq:z} to utilize only the mean estimate of the GP few-shot learner. While this utilizes the predictive power of the GP, we do not gain the full benefit of the additional uncertainty information. This already performs competitively with an mIoU of 63.1. We refer to this version as our baseline.

\parsection{Variance}Then, we add the variance as described by equation \eqref{eq:z}, leading to an impressive absolute gain of $3.3$ mIoU. This shows the benefit of incorporating uncertainty information from our GP. 

\parsection{Covariance}We further experiment with incorporating more advanced representations of the predictive covariance. Here, for each $z^j$ we collect the covariance of its spatial neighbours in a $5\times 5$ window. This leads to a minor performance increase of $0.44$.

\parsection{Mask Encoder}Last, we add the mask encoder proposed in~\cite{bhat2020learning}. This enables our GP to make predictions on a richer output space. This addition gives us an absolute mIoU gain of $1.3$. We therefore adopt this configuration as our final model.

\begin{table}[]
    \centering
    \begin{tabular}{c|c}
        \thickhline 
        Configuration & mIoU \\
        \thickhline
        Mean (Baseline) & 63.1\\
        Mean + Variance & 66.4\\
        Mean + Covariance & 66.8\\
        Mean + Covariance + Mask Encoder (\textbf{Ours})& 68.1\\
    \thickhline
    \end{tabular}
    \caption{Performance for different versions of our approach on the PASCAL-$5^i$ benchmark. Most notably, adding uncertainty information, in the form of variance or covariance, provides a significant gain in performance.}
    \label{tab:ablation}
\end{table}

\subsection{Choice of Kernel}
\label{sec:kernel} 

\begin{table}[]
    \centering
    \begin{tabular}{c|c}
        \thickhline 
        Method & mIoU \\
        \thickhline
        Linear & 65.1\\
        RQ & 68.1\\
        SE (\textbf{Ours}) & 68.1\\
    \thickhline
    \end{tabular}
    \caption{Comparison between different covariance functions $\kappa$ on the PASCAL-$5^i$ benchmark. The performance is stable between the squared exponential and rational quadratic kernels, while it significantly decreases for the linear kernel.}
    \label{tab:kernel_ablation}
\end{table}
We experiment with different covariance functions $\kappa$. In particular we compare the performance of the squared exponential (SE), rational quadratic (RQ), and linear (Lin) kernels. As discussed previously we use $\ell^2=\sqrt{D}$ for all our experiments where applicable. We present our results in Table \ref{tab:kernel_ablation}. Overall, the Gaussian process few-shot learner generalizes well to different kernel functions $\kappa$. This is perhaps not very surprising as the feature space on which the kernel act is learnt during training.

\parsection{Squared Exponential}
The squared exponential kernel is defined as
\begin{equation}
\label{eq:SE}
    \kappa_{SE}(x,y) = \exp\left(-\frac{||x-y||_2^2}{2\ell^2}\right)\enspace.
\end{equation}
This is equivalent to the full method discussed in our ablative analysis and achieves 68.1 mIoU. 

\parsection{Rational Quadratic}
The rational quadratic kernel is defined as
\begin{equation}
\label{eq:RQ}
    \kappa_{RQ}(x,y) = \left(1+\frac{||x-y||_2^2}{2\alpha\ell^2}\right)^{-\alpha}\enspace.
\end{equation}
In our experiments we use $\alpha=1$. We found that it achieves similar performance to the squared exponential with an mIoU of 68.1.

\parsection{Linear}
The linear kernel is defined as
\begin{equation}
    \label{eq:lin}
    \kappa_{Lin}(x,y) = x^Ty\enspace.
\end{equation}
Using a linear kernel leads to a significant performance decrease of $3.0$. The decrease in performance is in line with our previous reasoning since a Gaussian process with a linear kernel is equivalent to Bayesian linear regression \cite{Rasmussen2006}.

\section{Conclusion}

We have proposed a few-shot learner based on Gaussian process regression for the few-shot segmentation task. The GP models the support set in deep feature space and its flexibility permits it to capture complex feature distributions. It makes probabilistic predictions on the query image, providing both a point estimate and additional uncertainty information. These predictions are fed into a CNN decoder that predicts the final segmentation. The resulting approach obtains state-of-the-art performance on PASCAL-$5^i$ and COCO-$20^i$. It scales well with larger support sets during inference, even when trained for a fixed number of shots. With three or more shots, the approach outperforms the state-of-the-art on the COCO-$20^i$ benchmark.

\parsection{Acknowledgement} This work was partially supported by the Wallenberg AI, Autonomous Systems and Software Program (WASP) funded by the Knut and Alice Wallenberg Foundation; ELLIIT; and the ETH Z\"urich Fund (OK).

{\small
\bibliographystyle{ieee_fullname}
\bibliography{references}
}

\clearpage
\appendix
\begin{center}
  \section*{Appendix}
\end{center}
\renewcommand*{\thesection}{\Alph{section}}

In this appendix, we first provide additional details in Section~\ref{sec:appendixdetails}. Then, we supply additional results in Section~\ref{sec:results}. Last, we show additional qualitative examples and visualizations of the Gaussian process in Section~\ref{sec:qualitative}.

\section{Additional Implementation Details}\label{sec:appendixdetails}
We provide code for the Gaussian Process inference, the neural network layers that make up the modules used in our approach, and details on how we sample our episodes.

\subsection{Code for Gaussian Process}
Pseudo-code for the GP kernel is shown in Listing~\ref{lst:gpcode}. For brevity and clarity, we omit device casting and simplify the solve implementation. In practice, we use the standard triangular solver in PyTorch, \texttt{torch.triangular\_solve}.

\begin{listing}[b]
\begin{minted}{python}
def GP(x_q,y_s,x_s,sigma_y,kernel):
    """ Produces the predictive 
    posterior distribution of the GP
    
    Args: 
    x_q: deep query features (B,Q,D)
    y_s: support mask features (B,S,M)
    x_s: deep support features (B,S,D)
    sigma_y: mask standard deviation
    kernel: the kernel function,
    """
    B,S,D = x_s.shape
    I = torch.eye(S)
    K_ss = kernel(x_s,x_s) #(B,S,S)
    K_qq = kernel(x_q,x_q) #(B,Q,Q)
    K_sq = kernel(x_s,x_q) #(B,Q,S)
    L_ss = torch.cholesky(K_ss+
        sigma_y**2*I) #(B,S,S)
    mu_q = K_qs @ solve(L_ss.T,
        solve(L_ss,y_s)) #(B,Q,M)
    v = solve(L_ss,K_sq) #(B,S,Q)
    cov_q = K_qq - v.T @ v #(B,Q,Q)
    return mu_q,cov_q
\end{minted}
    \caption{PyTorch implementation of the Gaussian Process utilized in the proposed approach. Here, the learning and inference is combined in a single step. The \texttt{@} operator denotes matrix multiplication and \texttt{solve} the solving of a linear system of equations. The \texttt{.T} is the batched matrix transpose.}
    \label{lst:gpcode}
\end{listing}

\begin{table}[t]
    \centering
    \resizebox{\columnwidth}{!}{%
    \begin{tabular}{|l|l|r|}\hline
        \multicolumn{3}{|c|}{\textbf{Image Encoder}}\\\hline
        conv1   & \texttt{Conv2d}        & $64\times 224\times 224$\\
        bn1     & \texttt{BatchNorm2d}   & $64\times 224\times 224$\\
        relu1   & \texttt{ReLU}          & $64\times 224\times 224$\\
        maxpool & \texttt{MaxPool2d}     & $64\times 112\times 112$\\
        layer1  & \texttt{3x BottleNeck} & $256\times 112\times 112$\\
        layer2  & \texttt{4x BottleNeck} & $512\times 56\times 56$\\
        layer3  & \texttt{6x BottleNeck} & $1024\times 28\times 28$\\
        layer4  & \texttt{3x BottleNeck} & $2048\times 28\times 28$\\
        appearance & \texttt{Conv2d}     & $512\times 28\times 28$\\
        \hline
        \multicolumn{3}{|c|}{\textbf{Mask Encoder}}\\\hline
        conv1    & \texttt{Conv2d}      & $16\times 224\times 224$\\
        bn1      & \texttt{BatchNorm2d} & $16\times 224\times 224$\\
        relu1    & \texttt{ReLU}        & $16\times 224\times 224$\\
        maxpool  & \texttt{MaxPool2d}   & $16\times 112\times 112$\\
        layer1   & \texttt{BasicBlock}  & $32\times 56\times 56$\\
        layer2   & \texttt{BasicBlock}  & $64\times 28\times 28$\\
        conv2    & \texttt{Conv2d}      & $64\times 28\times 28$\\
        bn2      & \texttt{BatchNorm2d} & $64\times 28\times 28$\\
        relu2    & \texttt{ReLU}        & $64\times 28\times 28$\\
        \hline
        \multicolumn{3}{|c|}{\textbf{Decoder}}\\\hline
        conv1     & \texttt{Conv2d}      & $256\times 28\times 28$\\
        upsample1 & \texttt{Upsample}    & $256\times 56\times 56$\\
        cab1      & \texttt{CAB}         & $256\times 56\times 56$\\
        rrb1      & \texttt{RRB}         & $256\times 56\times 56$\\
        upsample2 & \texttt{Upsample}    & $256\times 112\times 112$\\
        cab2      & \texttt{CAB}         & $256\times 112\times 112$\\
        rrb2      & \texttt{RRB}         & $2\times 112\times 112$\\
        upsample3 & \texttt{Upsample}    & $2\times 448\times 448$\\
        \hline
    \end{tabular}}
    \caption{All neural network blocks used by our approach. The rightmost column shows the dimensions of the output of each block, assuming a $448\times 448$ input resolution. The image encoder is from He \etal~\cite{he2016deep}; the mask encoder from Bhat \etal~\cite{bhat2020learning}; and the decoder from Yu \etal~\cite{Yu2018dfn}. The \texttt{BottleNeck} and \texttt{BasicBlock} blocks are from He \etal~\cite{he2016deep}, and the \texttt{CAB} and \texttt{RRB} blocks from Yu \etal~\cite{Yu2018dfn}. See their works for additional details.}
    \label{tab:nnarch}
\end{table}

\begin{table*}[t]
    \centering
    \begin{tabular}{l | c c c c c | c c c c c }
        \thickhline
        \multirow{2}{*}{Method}          &        &        &1-Shot  &        &        &        &        & 5-Shot &        & \\
                          & F-0 & F-1 & F-2 & F-3 &  Mean  & F-0 & F-1 & F-2 & F-3 & Mean \\
        \thickhline
        RPMM~\cite{yang2020prototype}  & 36.3 & 55.0 & 52.5 & 54.6 & 49.6  & 40.2 & 58.0 & 55.2 & 61.8 & 53.8\\
        PFENet~\cite{Tian2020}         & 43.2 & 65.1 & \first{66.5} & 69.7 & 61.1  & 45.1 & 66.8 & 68.5 & 73.1 & 63.4\\
        RePRI~\cite{Boudiaf2020_repri} & \first{52.8} & 64.0 & 64.1 & 71.5 & \first{63.1}  & 57.7 & 66.1 & 67.6 & 73.1 & 66.2\\\hline
        Ours                           & 48.7 & \first{65.6} & 61.6 & \first{71.6} & 61.9  &  \first{65.8} & \first{71.7} & \first{71.9} & \first{80.2} & \first{72.4}\\
        \thickhline
    \end{tabular}
    \caption{The results of our approach in a COCO-$20^i$ to PASCAL transfer experiment (mIoU, higher is better). The approach is trained on a fold of COCO-$20^i$ training set and tested on the PASCAL validation set. The testing folds are constructed to include classes not present in the training set, and thus not the same as PASCAL-$5^i$.}
    \label{tab:domaintransfer}
\end{table*}

\begin{table*}[t]
    \centering
    \begin{tabular}{p{.23\textwidth} p{.23\textwidth} p{.23\textwidth} p{.23\textwidth}}\thickhline
        \textbf{Fold-0} & \textbf{Fold-1} & \textbf{Fold-2} & \textbf{Fold-3}\\\hline
        Airplane, Boat, Chair, Dining Table, Dog, Person & Bicycle, Bus, Horse, Sofa & Bird, Car, Potted Plant, Sheep, Train, TV-monitor & Bottle, Cat, Cow, Motorcycle\\\thickhline
    \end{tabular}
    \caption{The classes used for testing in the COCO-$20^i$ to PASCAL transfer experiment. This split is different from that of PASCAL-$5^i$ in order to avoid overlap between the training and testing classes.}
    \label{tab:cocotopascal}
\end{table*}

\subsection{Neural Network Architecture}
We supplement the neural network layers utilized in our approach in table~\ref{tab:nnarch}.

\subsection{Episode Sampling}
The way we sample episodes during training and evaluation follows prior work~\cite{Tian2020}. During training, we select a query image at random from the dataset, with the condition that it contains one of the classes considered. During evaluation, we go through the dataset images in sequence to select a query image. Next, the class is selected at random from one of the classes contained in the query image. Last, a support set is constructed. We make sure that (i) the support set does not contain the query images; and (ii) that the support set does not contain the same image more than once.

\section{Additional Results}\label{sec:results}
We supply three additional results. First, we show the results of a domain transfer experiment. Next, we provide the performance over different support set sizes also on PASCAL-$5^i$. Last, we report the runtimes of our approach.

\subsection{COCO-$20^i$ to PASCAL Transfer}
We show the effects of a domain transfer from COCO-$20^i$ to PASCAL in table~\ref{tab:domaintransfer}. This experiment follows the domain transfer experiment supplemented by Boudiaf \etal~\cite{Boudiaf2020_repri}. First, our approach is trained on each of the four folds of COCO-$20^i$. Next, we test each of the four versions on PASCAL, using only the classes held-out during training. We list the classes of each fold in table~\ref{tab:cocotopascal}. Our approach obtains a performance of 61.9 mIoU in the 1-shot setting and 72.4 mIoU in the 5-shot setting. Note that these results are not directly comparable to our result on the PASCAL-$5^i$ as the folds are not the same. However, the proposed approach obtains competitive performance in the 1-shot setting and sets a new state-of-the-art in the 5-shot setting. This is a clear example where the proposed approach manages to transfer between different datasets.

\subsection{Performance over Support Set Size}
We report the performance of the proposed approach on PASCAL-$5^i$ for different support set sizes. As in our original results, we use a model trained on 5 shots for all our results, except for the 1 shot case where we use a model trained on 1 shot. Additionally we compare the aggregated results to previous methods in Table~\ref{tab:numshots_pascal_aggr}.
\begin{table}[t]
    \centering
    \begin{tabular}{l |ccccc}\thickhline
        Num. Shots & Fold-0 & Fold-1 & Fold-2 & Fold-3 & \textbf{Mean}\\
        \thickhline
1-S&50.5 & 64.9& 54.6& 52.0& 55.5\\
2-S&57.5&64.8&55.5&50.4& 57.1\\
3-S&62.1&68.7&64.4&57.2& 63.1\\
4-S&64.7&70.2&69.4&59.8& 66.0\\
5-S&66.8&70.7&71.6&63.2& 68.1\\
6-S&67.2&71.5&72.3&64.1& 68.8\\
7-S&67.6&71.8&74.6&65.1& 69.8\\
8-S&67.9&72.9&75.0&66.2& 70.5\\
9-S&68.2&73.0&75.9&66.3& 70.8\\
10-S&68.4&73.5&76.5&67.0& 71.4\\
\thickhline
    \end{tabular}
    \caption{Peformance of our model when evaluated at different numbers of shots on the PASCAL-$5^i$ benchmark (mIoU, higher is better).}
    \label{tab:numshots_pascal}
\end{table}

\begin{table}[]
    \centering
    \begin{tabular}{l|ccc}
        \thickhline
         Method& 1-S & 5-S & 10-S  \\
         \thickhline
         RPMM~\cite{yang2020prototype}& 56.3 & 57.3 & 57.6\\
         PFENet~\cite{Tian2020}& \textbf{60.8} & 61.9 & 62.1\\
         RePRI~\cite{Boudiaf2020_repri}&59.7 & 66.6 & 68.1 \\
         \hline
         \textbf{Ours} & 55.5 & \textbf{68.1} & \textbf{71.4}\\
         \thickhline
    \end{tabular}
    \caption{Performance on PASCAL-$5^i$ for the 1-, 5- and 10-shot
settings averaged over 4 folds (mIoU, higher is better).}
    \label{tab:numshots_pascal_aggr}
\end{table}

\subsection{Runtimes}
We show the runtime of our method in Table~\ref{tab:runtimes}. We partition the timing into different parts. The Gaussian process (GP) is split into two. One part preparing the support set matrix and computing the matrix inverse in (4) and (5) of the main paper, and another part computing the mean and covariance given the pre-computed inverse.

The timings are measured in the 1-shot and 5-shot settings on images from COCO-$20^i$ of $512\times 512$ resolution, using a single episode at a time. We make use of the python-utility \texttt{time.perf\_counter()} and GPU-synchronization $torch.cuda.synchronize()$. We run our approach on a single NVIDIA V100 for 1000 episodes and report the average timings of each part. 
\begin{table}[t]
    \centering
    \begin{tabular}{l c c}
        \textbf{Function} & \textbf{1-shot time} & \textbf{5-shot time}\\\hline
        Image encoder on support  & 14 &  34\\
        Mask encoder on support   &  2 &   3\\
        GP preparation on support & 13 & 129\\
        Image encoder on query    & 12 &  14\\
        GP inference on query     & 11 &  12\\
        Decoder                   &  4 &   5\\\hline
        Total                     & 56 & 197\\\hline
    \end{tabular}
    \caption{Runtimes of the different functions in our approach, measured in milliseconds (ms). Timings are for a single episode, averaged over 1000 episodes, on $512\times 512$ size images from COCO-$20^i$.}
    \label{tab:runtimes}
\end{table}

\section{Qualitative Results}\label{sec:qualitative}
We provide additional qualitative results on COCO-$20^i$ and visualizations of the Gaussian process output on PASCAL-$5^i$.

\subsection{Additional Examples}
\begin{figure*}
    \hspace{.28\columnwidth}Support Set $\support$ \hspace{.28\columnwidth}\hspace{.03\columnwidth}Ground Truth Query \hspace{.03\columnwidth}\hspace{.10\columnwidth} Prediction
    
    \begin{overpic}[width=.4\columnwidth]{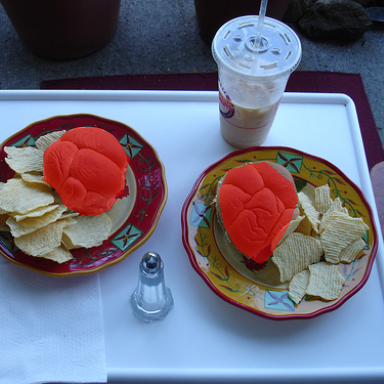}
    \put(100,0){\includegraphics[width=.2\columnwidth]{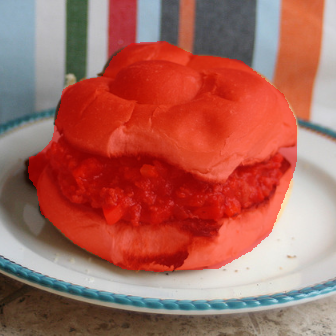}}
    \put(100,50){\includegraphics[width=.2\columnwidth]{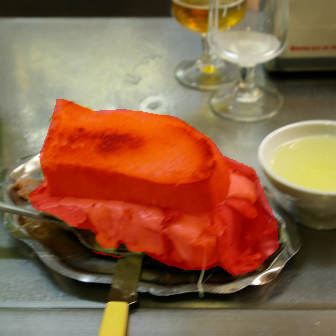}}
    \put(150,0){\includegraphics[width=.2\columnwidth]{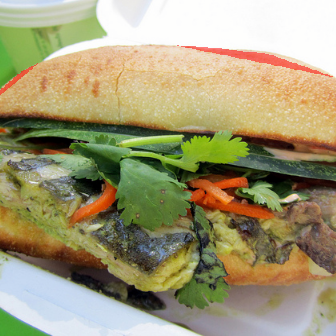}}
    \put(150,50){\includegraphics[width=.2\columnwidth]{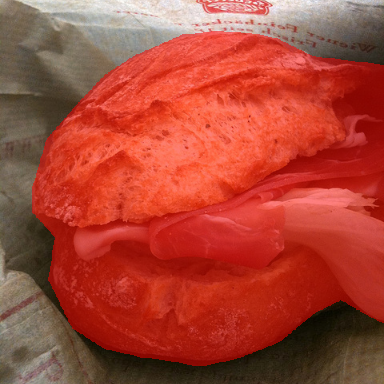}}
    \end{overpic}\hspace{.41\columnwidth}
    \includegraphics[width=.4\columnwidth]{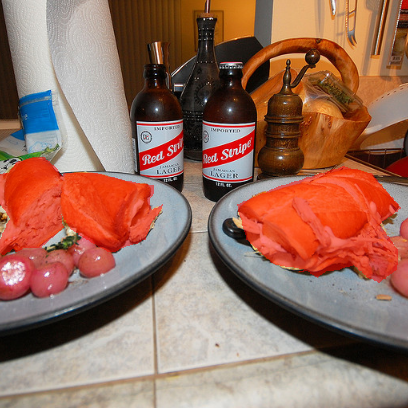}\hspace{0.01\columnwidth}
    \includegraphics[width=.4\columnwidth]{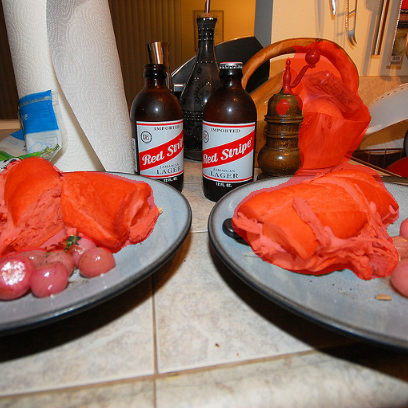}
    
    \vspace{0.01\columnwidth}
    \begin{overpic}[width=.4\columnwidth]{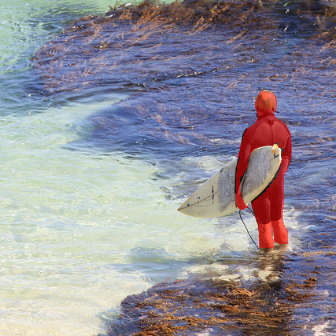}
    \put(100,0){\includegraphics[width=.2\columnwidth]{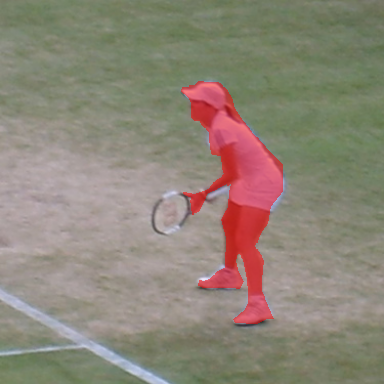}}
    \put(100,50){\includegraphics[width=.2\columnwidth]{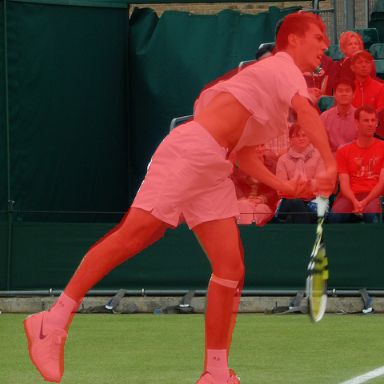}}
    \put(150,0){\includegraphics[width=.2\columnwidth]{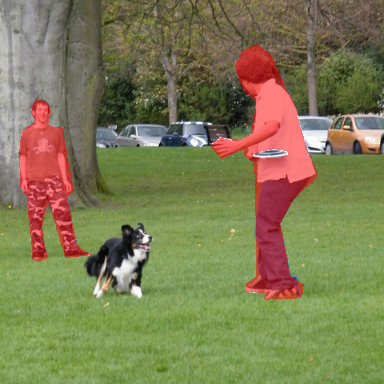}}
    \put(150,50){\includegraphics[width=.2\columnwidth]{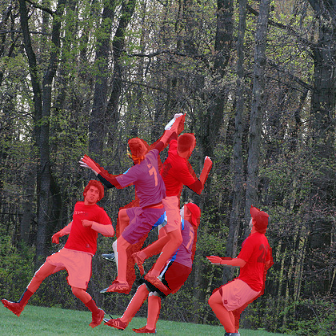}}
    \end{overpic}\hspace{.41\columnwidth}
    \includegraphics[width=.4\columnwidth]{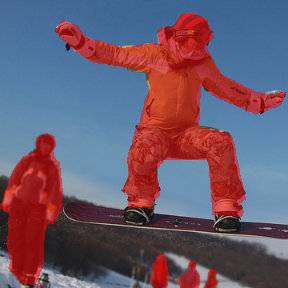}\hspace{0.01\columnwidth}
    \includegraphics[width=.4\columnwidth]{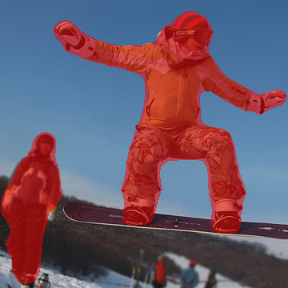}
    
    \vspace{0.01\columnwidth}
    \begin{overpic}[width=.4\columnwidth]{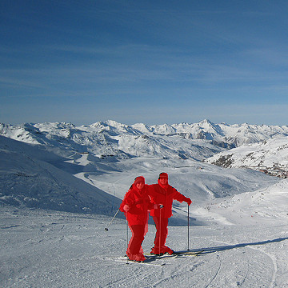}
    \put(100,0){\includegraphics[width=.2\columnwidth]{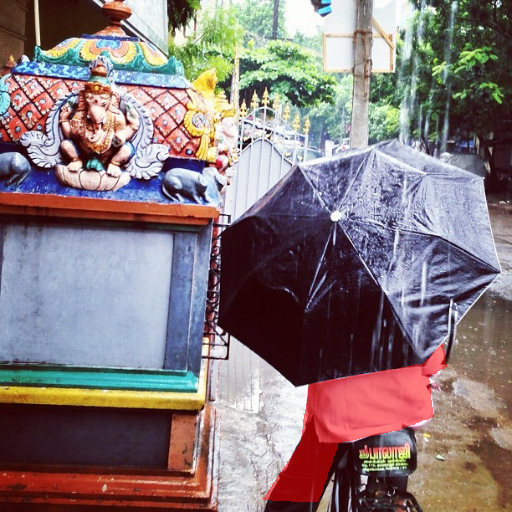}}
    \put(100,50){\includegraphics[width=.2\columnwidth]{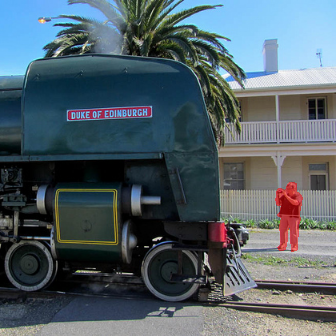}}
    \put(150,0){\includegraphics[width=.2\columnwidth]{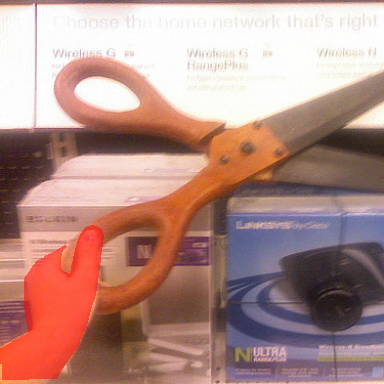}}
    \put(150,50){\includegraphics[width=.2\columnwidth]{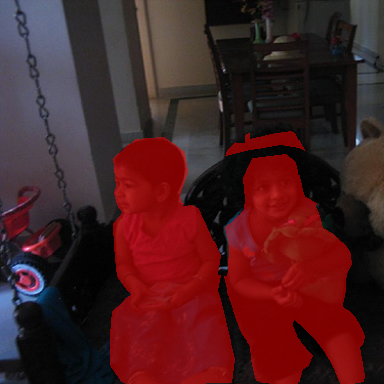}}
    \end{overpic}\hspace{.41\columnwidth}
    \includegraphics[width=.4\columnwidth]{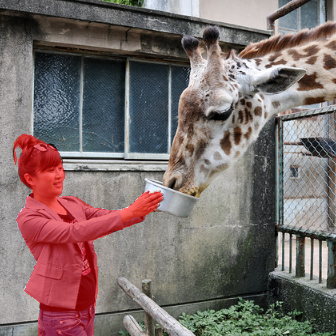}\hspace{0.01\columnwidth}
    \includegraphics[width=.4\columnwidth]{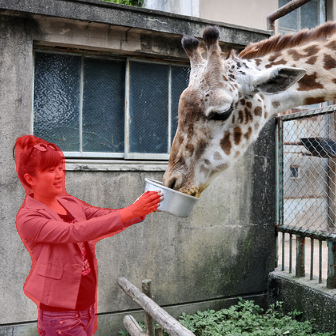}

    \vspace{0.01\columnwidth}
    \begin{overpic}[width=.4\columnwidth]{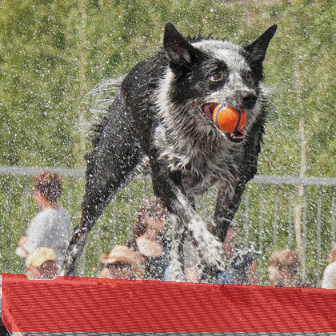}
    \put(100,0){\includegraphics[width=.2\columnwidth]{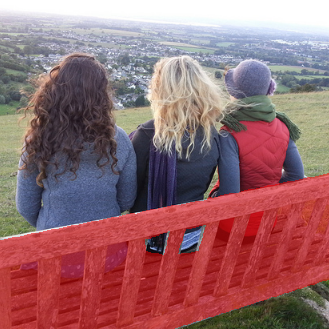}}
    \put(100,50){\includegraphics[width=.2\columnwidth]{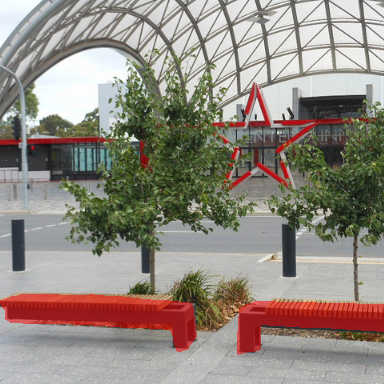}}
    \put(150,0){\includegraphics[width=.2\columnwidth]{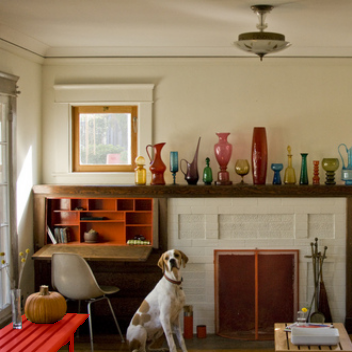}}
    \put(150,50){\includegraphics[width=.2\columnwidth]{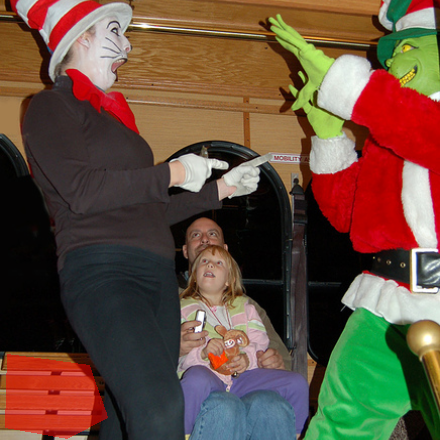}}
    \end{overpic}\hspace{.41\columnwidth}
    \includegraphics[width=.4\columnwidth]{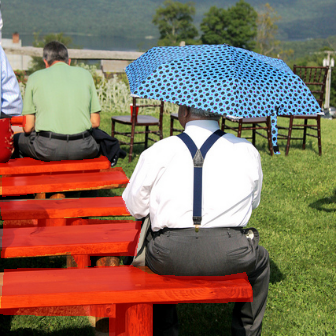}\hspace{0.01\columnwidth}
    \includegraphics[width=.4\columnwidth]{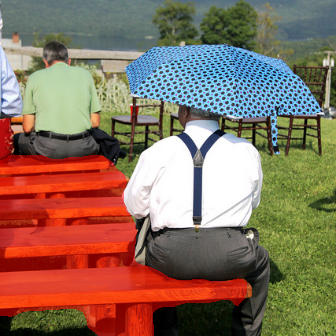}

    \vspace{0.01\columnwidth}
    \begin{overpic}[width=.4\columnwidth]{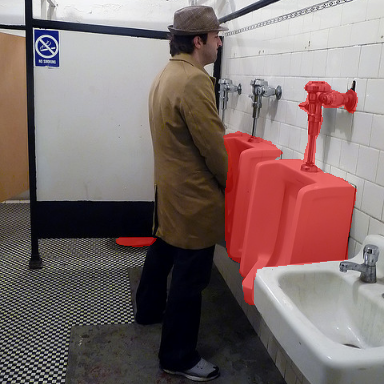}
    \put(100,0){\includegraphics[width=.2\columnwidth]{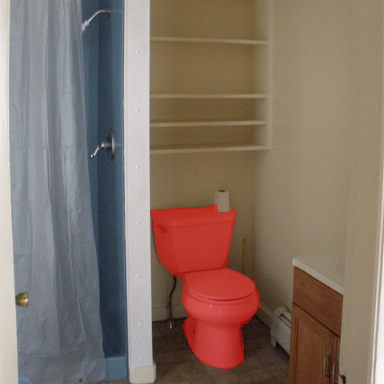}}
    \put(100,50){\includegraphics[width=.2\columnwidth]{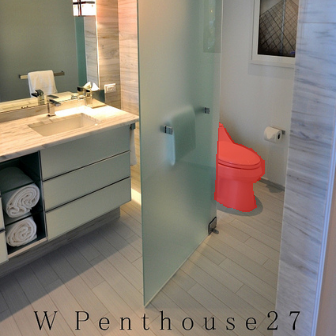}}
    \put(150,0){\includegraphics[width=.2\columnwidth]{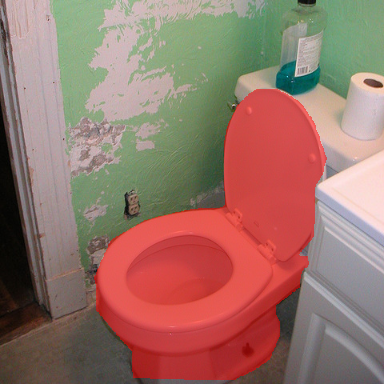}}
    \put(150,50){\includegraphics[width=.2\columnwidth]{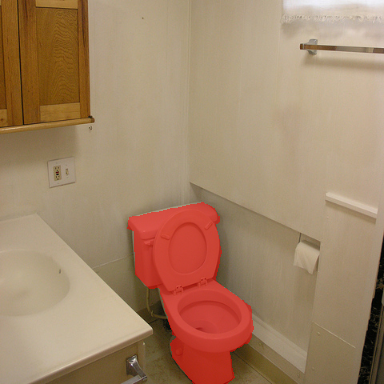}}
    \end{overpic}\hspace{.41\columnwidth}
    \includegraphics[width=.4\columnwidth]{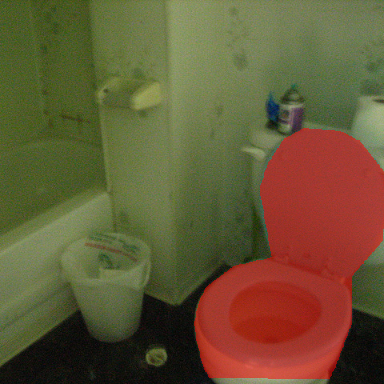}\hspace{0.01\columnwidth}
    \includegraphics[width=.4\columnwidth]{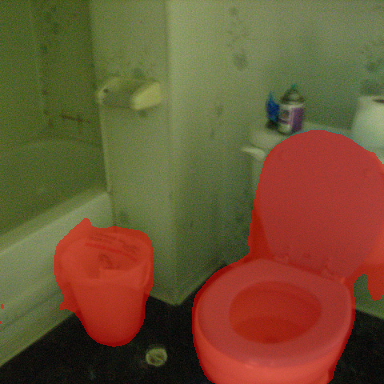}

    \vspace{0.01\columnwidth}
    \begin{overpic}[width=.4\columnwidth]{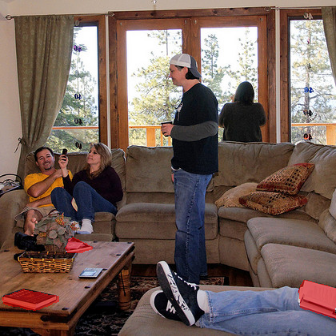}
    \put(100,0){\includegraphics[width=.2\columnwidth]{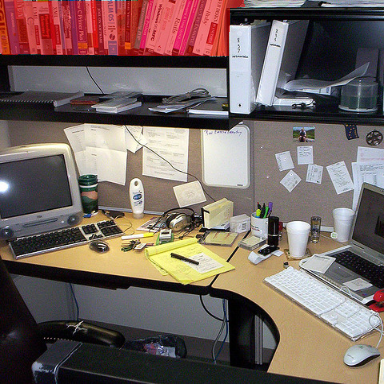}}
    \put(100,50){\includegraphics[width=.2\columnwidth]{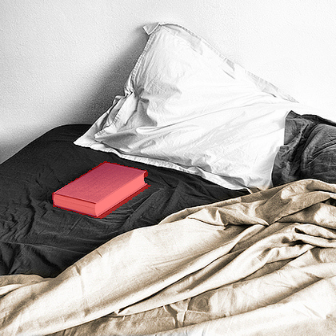}}
    \put(150,0){\includegraphics[width=.2\columnwidth]{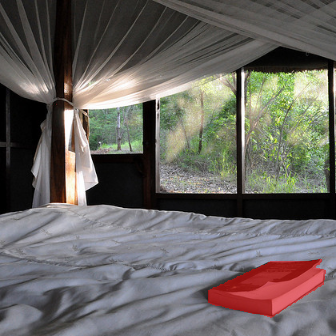}}
    \put(150,50){\includegraphics[width=.2\columnwidth]{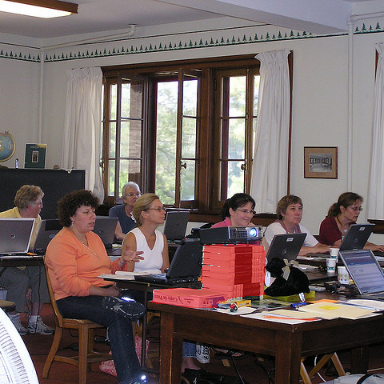}}
    \end{overpic}\hspace{.41\columnwidth}
    \includegraphics[width=.4\columnwidth]{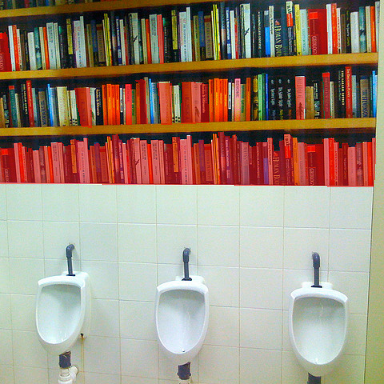}\hspace{0.01\columnwidth}
    \includegraphics[width=.4\columnwidth]{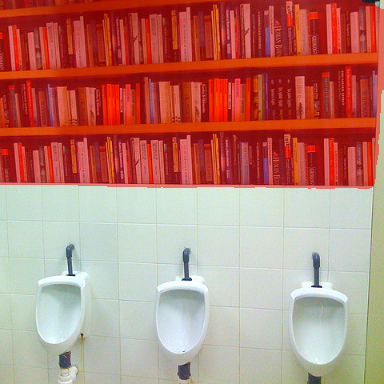}

    \vspace{0.01\columnwidth}

\caption{Additional qualitative results of our approach on COCO-$20^i$. We show the five support images with annotations overlaid (left), the query image and annotation (center), and predictions made by our model (right). Note that the images are cropped for aesthetic reasons.}
\label{fig:appendixqualitative}
\end{figure*}

In Figure~\ref{fig:appendixqualitative} we show qualitative results on COCO-$20^i$. Note that in the last row, the annotation is not complete. Only one out of three rows of books placed above some urinals are annotated. Our approach instead correctly segments out all three rows of books.

\subsection{Visualization of Gaussian Process Output}
\begin{figure*}
    \hspace{0.0\columnwidth}Support~Set \hspace{.33\columnwidth}Query \hspace{.36\columnwidth}Prediction \hspace{.33\columnwidth}Gaussian process
    \centering
    \includegraphics[width=\textwidth]{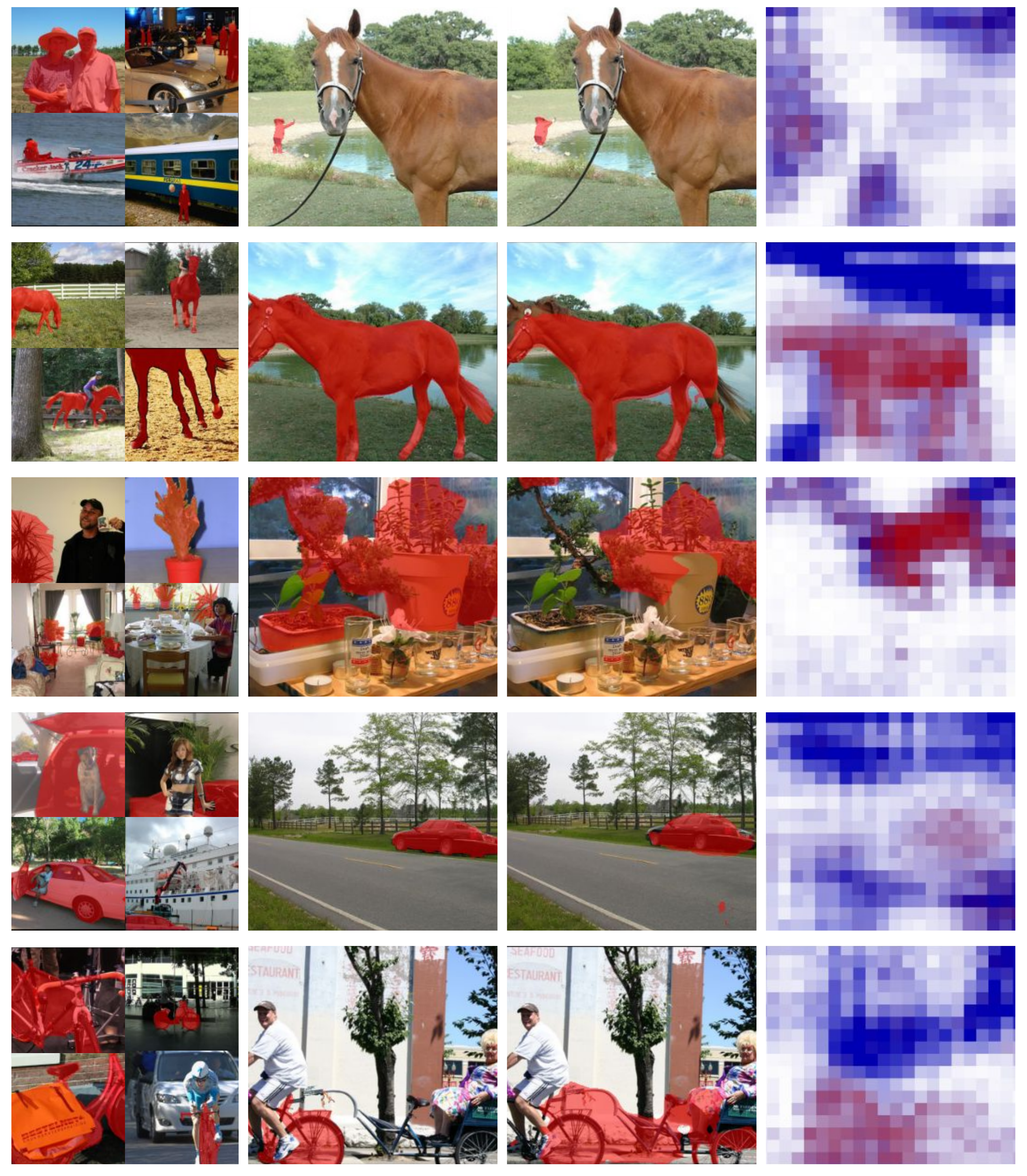}
    \caption{Visualization of the Gaussian process module output on PASCAL-$5^i$. Here, we use a version of our approach (\emph{Mean + Covariance} in Table~3 of the main paper) with a simple mask encoder to increase interpretability of the Gaussian process. The approach is given four support images. We show the mean $\mu_{\query|\support}$ in red-blue. Areas with high variance are shown in white. Images have been cropped for aesthetic reasons.}
    \label{fig:gpout}
\end{figure*}

We provide a visualization of the Gaussian process output in Figure~\ref{fig:gpout}. In the top row, the approach correctly segments the person class. The background in this example contains a large horse. There are no horses in the support set and the Gaussian process is uncertain in that region, reporting high variance. In the second row, we instead aim to segment the horse-class. The support set contains multiple horses similar to that in the query image, and the Gaussian process accurately reports the horse as foreground with low variance. In the third row, our approach misses parts of the plants. These parts are marked with high variance, despite fairly similar examples being present in the support set. In the fourth row, our approach segments a small part of the background, but otherwise performs well. In the fifth row, a carriage is deemed similar to a bicycle and segmented as such. The Gaussian process reports a high variance for the persons but a low variance for large parts of the carriage.

\end{document}